\documentclass[]{article}

\pdfoutput=1

\usepackage[preprint]{neurips_2023}
\usepackage{amsmath}

\usepackage[utf8]{inputenc} 
\usepackage[T1]{fontenc}    

\usepackage{url}            
\usepackage{booktabs}       
\usepackage{amsfonts}       
\usepackage{graphicx}
\usepackage{nicefrac}       
\usepackage{microtype}      
\usepackage{xcolor}         
\usepackage{bm}             
\usepackage{subcaption}
\usepackage{natbib}
\setcitestyle{authoryear,open={(},close={)}} 

\usepackage{paralist}

\begin{document}

\title{Identifying Shared Decodable Concepts
 in the Human Brain Using Image-Language Foundation Models}

%

\author{Cory Efird \\
    Computing Science and Psychology\\
    University of Alberta \\
    \texttt{efird@ualberta.ca} \\
    \And
    Alex Murphy \\
    Computing Science and Psychology\\
    University of Alberta \\
    \texttt{amurphy3@ualberta.ca} \\
    \AND
    Joel Zylberberg \\
    Physics and Astronomy \\
    York University \\
    \texttt{joelzy@yorku.ca} \\
    \And
    Alona Fyshe \\
    Computer Science and Psychology\\
    University of Alberta \\
    \texttt{alona@ualberta.ca} \\
}

\maketitle

\begin{abstract}

We introduce a method that takes advantage of high-quality pretrained multimodal representations to explore fine-grained semantic networks in the human brain. Previous studies have documented evidence of functional localization in the brain, with different anatomical regions preferentially activating for different types of sensory input. Many such localized structures are known, including the fusiform face area and parahippocampal place area. This raises the question of whether additional brain regions (or conjunctions of brain regions) are also specialized for other important semantic concepts. To identify such brain regions, we developed a data-driven approach to uncover visual concepts that are decodable from a massive functional magnetic resonance imaging (fMRI) dataset. Our analysis is broadly split into three sections. First, a fully connected neural network is trained to map brain responses to the outputs of an image-language foundation model, CLIP \citep{CLIP}. Subsequently, a contrastive-learning dimensionality reduction method reveals the brain-decodable components of CLIP space. In the final section of our analysis, we localize shared decodable concepts in the brain using a voxel-masking optimization method to produce a shared decodable concept (SDC) space. The accuracy of our procedure is validated by comparing it to previous localization experiments that identify regions for faces, bodies, and places. In addition to these concepts, whose corresponding brain regions were already known, we localize novel concept representations which are shared across participants to other areas of the human brain. We also demonstrate how this method can be used to inspect fine-grained semantic networks for individual participants. We envisage that this extensible method can also be adapted to explore other questions at the intersection of AI and neuroscience.

\end{abstract}

\section{Introduction}


To navigate the world, individuals must learn to quickly interpret what they see.  Evolution created pressure to quickly extract certain types of visual information.  For example recognizing and interpreting faces is core to many of our social interactions, recognizing animate objects is key to avoiding a predator (or pursuing prey). 
As a byproduct, the visual system identifies a core set of concepts necessary for a successful existence in our world.   But what are these core concepts, and how does the brain represent them?  Seeking the answer to this question has been central to decades of neuroscience research. 

Some have argued that the brain has specific areas tuned to detecting specific concepts.  There is significant evidence suggesting there are areas of the brain that preferentially activate for stimuli containing faces~\citep{Kanwisher1997}, places~\citep{Epstein1998}, and more recently, there have been reports of food-specific brain areas~\citep{jain2023selectivity, Pennock2023}.  The controversy around these findings is driven largely by the observation that the brain areas are not ``tuned'' specifically for faces or places; they also respond to other visual stimuli, meaning they are not face- or place-specific~\citep{haxby_2001, Hanson2008}.  Thus, the key question remains unanswered: Are there dimensions of meaning recoverable from the brain's responses to image stimuli that are consistent in 1) content, and 2) localization across participants?  

In this work we take an entirely data-driven approach to uncovering dimensions of meaning within the human brain.  We use the Natural Scenes Dataset \citep{NSD}, one of the largest and most comprehensive visual stimulus functional Magnetic Resonance Imaging (fMRI) datasets to date, and CLIP, a shared text and image embedding space~\citep{CLIP}.  We present a new decoding model that produces high top-1 accuracy, predicting CLIP space from fMRI.  We then use the \emph{predicted} CLIP space to learn a new embedding space we call the Shared Decodable Concept (SDC) space.  SDC-space: 
\begin{compactenum}
    \item is trained across participants specifically to identify the dimensions of meaning that are decodable from fMRI
    \item has a small number of coherent concepts per dimension
    \item shows consistent cross-participant localization in brain-space
\end{compactenum}
SDC-space allows for a data-driven mapping of concepts to brain areas, which allowed us to find several new concepts localized to specific brain areas.

The hunt for concept-specific areas of the brain has been a decades-long venture.  Very early work focused on the tuning of neurons for very low level features~\cite{Hubel1959}, followed by the discovery of brain areas preferentially activated for specific concepts and dimensions of semantic meaning ~\citep{Kanwisher1997, McCandliss2003, Uta_tools}. These studies were largely hypothesis-driven, with stimuli chosen specifically to search for areas tuned to certain concepts.  

Hypothesis-driven experimental design is a cornerstone of neuroscience research, and has contributed greatly to our current understanding of the brain.  However, recently some have argued for a more data-driven naturalistic approach to neuroscience~\citep{Matusz2019, Hamilton2020, kanwisher2022food}.  Our work differs from previous data-driven approaches in that we perform dimensionality reduction in CLIP embedding space. For contrast, other work does the dimensionality reduction directly in fMRI voxel space, often for a single ROI. By using voxels from multiple ROIs, our method makes use of all of the information decodable from cortex.  In addition, or SDC space has dimensions that are highly interpretable. This is in stark contrast to other methods, like Principal Components Analysis (PCA) where the different dimensions often lack interpretability.  Our SDC space also leverages connections between the true and predicted CLIP space, which is not possible with typical PCA-style analyses.

A data-driven approach has the potential to uncover brain areas tuned to concepts that we might not otherwise have considered, as well as to expand our understanding of the specificity of certain brain areas beyond narrow visual concept classes.  What follows is a framework for uncovering such visual concepts that suggests multiple new avenues for future hypothesis-driven research.

\section{Decoding CLIP-Space from Brain Images}\label{Section1}
To identify shared decodable concepts in the brain, we require a mapping from brain space to a suitable representational space. In this section we describe the pieces necessary for creating this mapping: a multimodal image-language embedding model (CLIP), a brain imaging dataset (NSD), and our method to map from per-image brain responses to their associated multimodal embeddings. We consider two models in this section, and verify that our proposed neural network decoder outperforms a regression-based linear model.

\subsection{Data}

\paragraph{fMRI Data}
The natural scenes dataset (NSD) is a massive fMRI dataset acquired to study the underpinnings of natural human vision. Eight participants were presented with 30,000 images (10,000 unique images over 3 repetitions) from the Common Objects in Context (COCO) naturalistic image dataset \citep{coco}.  A set of 1,000 shared images were shown to all participants, while the other 9,000 images were unique to each participant. Single-trial beta weights were derived from the fMRI time series using the GLMSingle toolbox \citep{GLMsingle}. This method fits numerous haemodynamic response functions (HRFs) to each voxel, as well as an optimised denoising technique and voxelwise fractional ridge regression, specifically optimised for single-trial fMRI acquisition paradigms. Some participants did not complete all sessions, and three sessions were held out by the NSD team for the Algonauts challenge. Further details can be found in \cite{NSD}.



\paragraph{Representational Space for Visual Stimuli}
To generate representations for each stimulus image, we use a model trained on over 400 million text-image pairs with the contrastive language-image pretraining objective (CLIP~\citep{CLIP}). The CLIP model consists of a text-encoder and image-encoder that are jointly trained to maximize the cosine similarity of corresponding text and image embeddings in a shared low-dimensional space. We use the 32-bit Transformer model (ViT-B/32) implementation of CLIP to create 512-dimensional representations for each of the stimulus images used in the NSD experiment. We train our decoder to map from fMRI responses during image viewing to the associated CLIP vector for that same image.

\subsection{Data Preparation}\label{sec:data_prep}

\paragraph{Data Split}
We randomly split the per-image brain responses $\bm{X}$ and CLIP embeddings $\bm{Y}_{\text{CLIP}}$ into training $(\bm{X}^{\text{Train}}, \bm{Y}^{\text{Train}}_{\text{CLIP}})$, validation $(\bm{X}^{\text{Val}}, \bm{Y}^{\text{Val}}_{\text{CLIP}})$, and test $(\bm{X}^{\text{Test}}, \bm{Y}^{\text{Test}}_{\text{CLIP}})$ folds. The validation and test folds were each chosen to have exactly 1,000 images. Some participants in the NSD did not complete all scanning sessions and only viewed certain images once or twice. We assign these images to the training set. Of the shared 1,000 images, 413 were shown three times to every participant across the sessions released by NSD. These 413 images appear in each participant's testing fold. 

\paragraph{Voxel Selection}
The noise ceiling is often used to identify the voxels that most reliably respond to visual stimuli. The NSD fMRI data comes with voxelwise noise ceiling estimates, but they are calculated using the \emph{full dataset}. These estimates can be used to extract a subset of voxels for decoding analyses, but this takes into account voxel sensitivity to images we later want to tune and test on, and is a form of double-dipping \citep{doubledipping}. We therefore re-calculated the per-voxel noise ceiling estimates specifically on the designated training data only. Voxels with noise ceiling estimates above $5\%$ variance explainable by the stimulus were used as inputs to the decoding model, resulting in  10k-30k voxel subsets per participant (see Supplementary Info for exact per-participant voxel dimensions).

\subsection{Decoding Methodology}

\begin{figure}
    \centering
    \begin{subfigure}[b]{.64\textwidth}
        \centering
        \includegraphics[width=9cm]{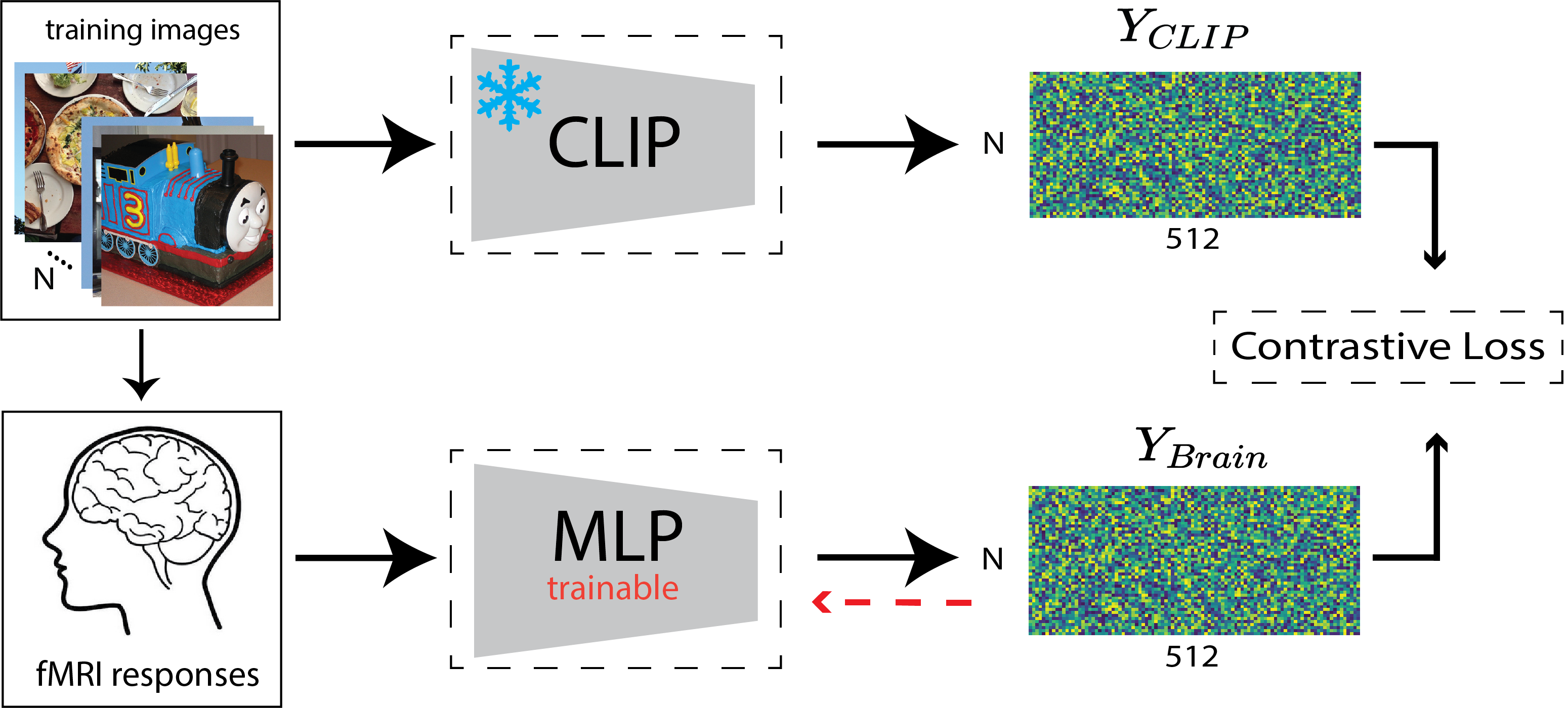}
        \caption{Decoding CLIP representations from the brain. We create CLIP representations by passing NSD training set stimuli to CLIP (frozen).  The NSD team showed these same images to human participants during fMRI scans. We train a single-layer MLP to predict CLIP embeddings from the fMRI responses to the corresponding images. We update the MLP weights using a contrastive loss function (Eq.~\ref{eq:contrastive_loss}). Black arrows represent the flow of data through the procedure and dashed red lines represent gradient updates used to train the model.}
        \label{fig:sub1}
    \end{subfigure}
    \hspace{2mm}
    \begin{subfigure}[b]{.32\textwidth}
        \centering
        \includegraphics[width=3.5cm]{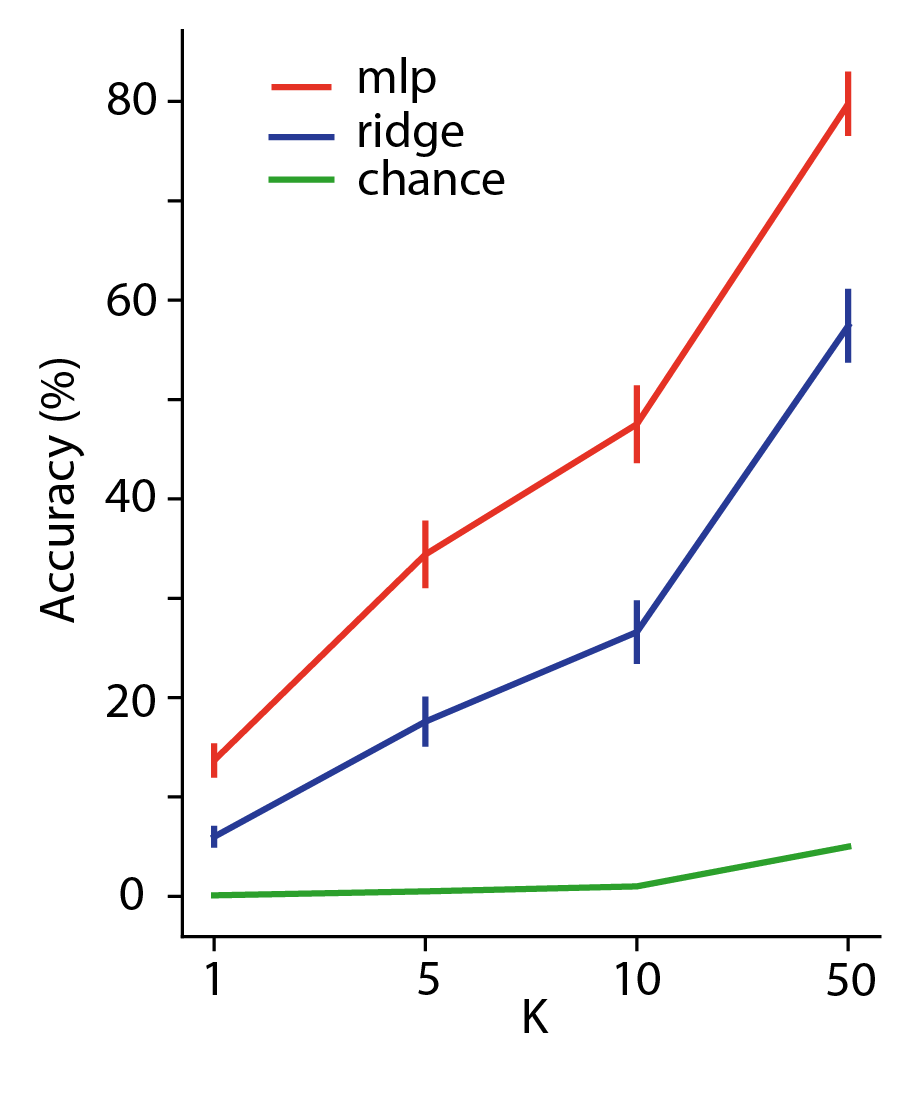}
        \caption{Top-k accuracy for CLIP decoding using the proposed MLP and a ridge regression baseline. The MLP implementation outperforms the ridge regression baseline across various values of $k$. Chance performance is given in green. Accuracy calculated on test data.}
        \label{fig:sub2}
    \end{subfigure}
    \label{fig:clip_decode}
    \centering
    \caption{(a) The training paradigm for deriving brain-decoded CLIP embeddings. (b) Evaluation results of our model with respect to a regression baseline.}
\end{figure}

\paragraph{Decoding Model}
The decoding model $\bm{g}: \mathbb{R}^{v} \to \mathbb{R}^{512}$ is trained to map a vector of brain responses $\bm{X} = [\bm{x}_1, \ldots , \bm{x}_n], \bm{x}_i \in \mathbb{R}^{v}$ to the CLIP embeddings of the corresponding stimulus images $\bm{Y}_{\text{CLIP}} = [\bm{y}_1, \ldots, \bm{y}_n], \bm{y}_i \in 
\mathbb{R}^{512}$.   Here $n$ is the number of training instances, and $v$ is the number of voxels. An illustration of the decoding procedure is given in \ref{fig:sub1}. We define $\bm{g}$ to be a multi-layer perceptron (MLP) with a single hidden layer of size 5,000, following by a leaky ReLU activation (slope=0.01). The model is trained for 12 epochs (approximately 1,500 iterations) with the Adam optimizer and a batch size of 128. The learning rate is initialized to $1e^{-4}$ and decreased by a factor of 10 after epochs 3, 6, and 9. We train the brain decoder using the the InfoNCE loss function \citep{infoNCE}, which is defined as:

\begin{equation}\label{eq:contrastive_loss}
    \mathcal{L}_{\text{contrastive}} = -log \frac{exp(q\cdot k_{+}/\tau)}{\sum_{i=0}^{K}exp(q\cdot k_{i}/\tau)}
\end{equation}

In the original CLIP setting, $q$ and $k$ both represent image and language embeddings, where the loss is minimized when these embeddings have a high similarity for the same images (positive class) and low similarity otherwise. In our implementation, we replace the language embeddings with the brain responses to images. We set $\tau = 1$ in our implementation. We compare our model $\bm{g}$ to a baseline ridge regression model trained on the same data. We used grid search to select the best ridge regularization parameter $\lambda \in \{0.1, 1, 10, 100, 1000, 10000, 100000\}$ using the validation data.

\paragraph{Evaluation}
We evaluate our models using top-k accuracy, which is computed by sorting in ascending order all true representations $\{y_1 \ldots y_n \}$ by their cosine distance to a predicted representation $\hat{y}_i$. Top-k Accuracy is the percentage of instances for which the true representation $y_i$ is amongst the top-k items in the sorted list. Chance top-k accuracy is $\frac{100 \cdot k}{n} \%$ where $n$ is the number of held-out data points used for evaluation.
Figure \ref{fig:sub2} shows the results of this evaluation. The MLP model outperforms ridge regression across all values of $k$, motivating the need for this more complex model.

Recall that our end goal is to identify shared decodable concepts (SDC) in the brain.  Our methodology for this task relies on the \emph{predicted} CLIP vectors, and so an accurate deocoding model is of utmost importance. 



\section{Optimizing for Shared Decodable Concepts by Transforming CLIP-space}\label{optimising-concepts}

Because CLIP was trained on images and text, it is an efficient embedding space for those modalities.  However, we are interested in the dimensions of meaning available in images that are \emph{decodable from fMRI recordings}. To explore the brain-decodable dimensions of meaning in CLIP space we pursued two directions.  First, we learned a mapping to transform CLIP into a pre-existing 49-dimensional model which was trained to on human behavioral responses to naturalistic image data from the THINGS-initiative \citep{THINGS}.  Second, to find dimensions of meaning specifically available in fMRI data, we explored several possible linear projections of the brain-decoded CLIP embeddings, $\bm{Y}_{\text{Brain}}$. This method, which combines predicted CLIP embeddings across participants, produced dramatic increases in top-1 accuracy over the original CLIP space and the THINGS space transformation of CLIP (Figure \ref{fig:sub2})

\subsection{THINGS Concepts}

\paragraph{THINGS Experiment}
The THINGS-Images database consists of 1854 object classes with 12 images per class \citep{THINGS}. These images were used to gather approximately 1.46 million responses to an odd-one-out task, during which MTurk workers were asked to select the one image (from a group of 3) that least belonged in the group. These responses were then used to learn THINGS object embeddings $[\bm{t}_1,..., \bm{t}_{1854}] = \bm{T}$ , $\bm{t}_i \in \mathbb{R}^{49}$. Initially, $\bm{T}$ is randomly initialized. Then, for each triplet of classes $i, j, k$, and the human-chosen odd one out ($i$), the model is trained to maximize the dot products of the embeddings for the non-odd-one-out concepts ($\bm{t}_j \cdot \bm{t}_k$), and minimize the other two dot products ($\bm{t}_i \cdot \bm{t}_j$ and $ \bm{t}_i \cdot \bm{t}_k$). The model is constrained ensure non-negativity and encourage sparsity in $\bm{T}$.  After training, $\bm{T}$ contains 49 human-interpretable dimensions that are most important for performing the odd-one-out similarity judgments. These 49 dimensions were assigned semantic labels by hand.

\begin{figure}
    \centering

    \includegraphics[width=\textwidth]{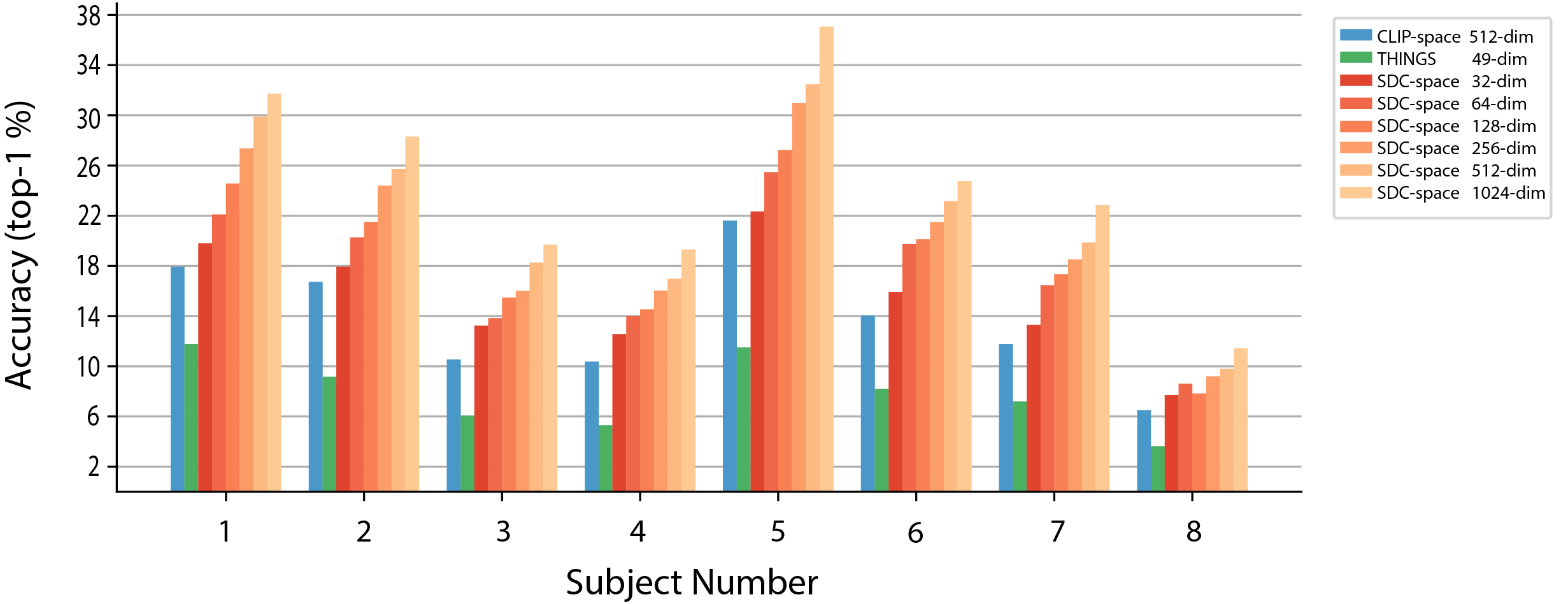}
    \caption{We can improve brain-decoded CLIP embeddings by optimising a transformation on the output of our MLP decoder (blue bar) outlined in Section \ref{Section1}. Transforming to THINGS-space (green bar) results in a drop in top-1 accuracy over our initial decoder. We find that by optimizing a specialized projection using data pooled across participants, $W_{SDC}$, we can dramatically improve brain-decoding to CLIP space.}
    \label{fig:spaces}

\end{figure}

\paragraph{Translation from CLIP to THINGS space}
We trained a function $\bm{h}_{\text{THINGS}}: \mathbb{R}^{512} \to \mathbb{R}^{49}$  to map from CLIP space to THINGS space. To train this mapping, all 12 images for each of the 1,854 Things object classes are passed into the CLIP image encoder to obtain $\bm{U} \in  \mathbb{R}^{(12\cdot1845) \times 512}$. We then averaged CLIP embeddings for all images within an object class to obtain $\bm{U}_{\text{avg}} \in \mathbb{R}^{ 1854 \times 512}$. These averaged embeddings are used to fit a linear ridge regression model:
\begin{equation}\label{eq:THINGS}
    \min_{\bm{W}_\text{THINGS}} \quad
    \lVert \bm{U}_{\text{avg}}\bm{W}_\text{THINGS}^T - \bm{T}\rVert_2^2 + 
    \alpha \lVert \bm{W}_\text{THINGS} \rVert_2^2
\end{equation}  
Because the THINGS-embeddings $\bm{T}$ are constrained to be non-negative, after training we introduced an additional ReLU operation, $\bm{h}_{\text{THINGS}}(\bm{Y}) = \text{ReLU}(\bm{Y}\bm{W}_{\text{THINGS}}^T)$, and further finetuned $\bm{W}_{\text{THINGS}}$.  Empirical results for this fine tuning can be seen in the Supplementary Material. The CLIP-to-THINGS mapping function $\bm{h}_{\text{THINGS}}$ allows us to map CLIP embeddings derived from NSD stimulus images to THINGS space.  In addition, we can use brain responses to those images passed through the decoder $\bm{g}(\bm{X})$ to derive decoded THINGS embeddings directly from the NSD fMRI.

Results for computing top-k accuracy in THINGS space appear in Figure~\ref{fig:spaces} (green bar).  Decoding performance in THINGS space is lower than in the original CLIP space.  This implies that the THINGS concepts, originally derived from behavioral data, do not sufficiently capture the visual concepts available in CLIP space that are decodable from fMRI.  This motivates the search for a new embedding space tuned specifically for the decodable concepts shared amongst all participants in the NSD dataset.

\subsection{Finding Shared Decodable Concepts (SDC)}\label{finding_SDCs}

In this section, we describe our method for deriving a transformation of CLIP space that reveals brain-decodable interpretable dimensions that are shared across participants. First, each participant's brain decoded and ground truth stimulus representations $\bm{Y}_{\text{CLIP}}^{\text{Val}}, \bm{Y}_{\text{Brain}}^{\text{Val}} \in \mathbb{R}^{3000 \times 512}$ are concatenated into shared matrices $\bm{Y}_{\text{CLIP}}^{\text{ValAll}}, \bm{Y}_{\text{Brain}}^{\text{ValAll}} \in \mathbb{R}^{8\cdot3000 \times 512}$. The function that maps CLIP space to a shared decodable concept space (SDC)  $\bm{h}_{\text{SDC}}: \mathbb{R}^{512} \to \mathbb{R}^c$ where $c$ is the chosen dimensionality of the SDC-space. Similar to $\bm{h}_{\text{THINGS}}$, the mapping function is defined as a multiplication by a weight matrix followed by a ReLU, i.e. $\bm{h}_{\text{SDC}}(\bm{Y}) = \text{ReLU}(\bm{Y} \bm{W}_\text{SDC}^T)$. The weight matrix $\bm{W}_\text{SDC}$ is found by optimizing
\begin{equation}
    \min_{\bm{W}_\text{SDC}} \quad
    \mathcal{L}_{\text{contrastive}} 
    (
        \text{LeakyReLU}(\bm{Y}_{\text{CLIP}}^{\text{Val}}  \bm{W}_\text{SDC}^T) , 
        \text{LeakyReLU}(\bm{Y}_{\text{Brain}}^{\text{Val}} \bm{W}_\text{SDC}^T) 
    )
\end{equation}  
The weight matrix $\bm{W}_\text{SDC}$ is randomly initialized and trained for 10,000 iterations using the Adam optimizer, a batch size of 3,000, learning rate 2e-4. A leaky ReLU with a negative slope of 0.05 is used during optimization because it encourages convergence of the SDC space components. The SDC model is fit for different numbers of components $c \in \{32, 64, 128, 256, 512, 1024\}$.

\begin{figure}
    \centering
    \includegraphics[width=13cm]{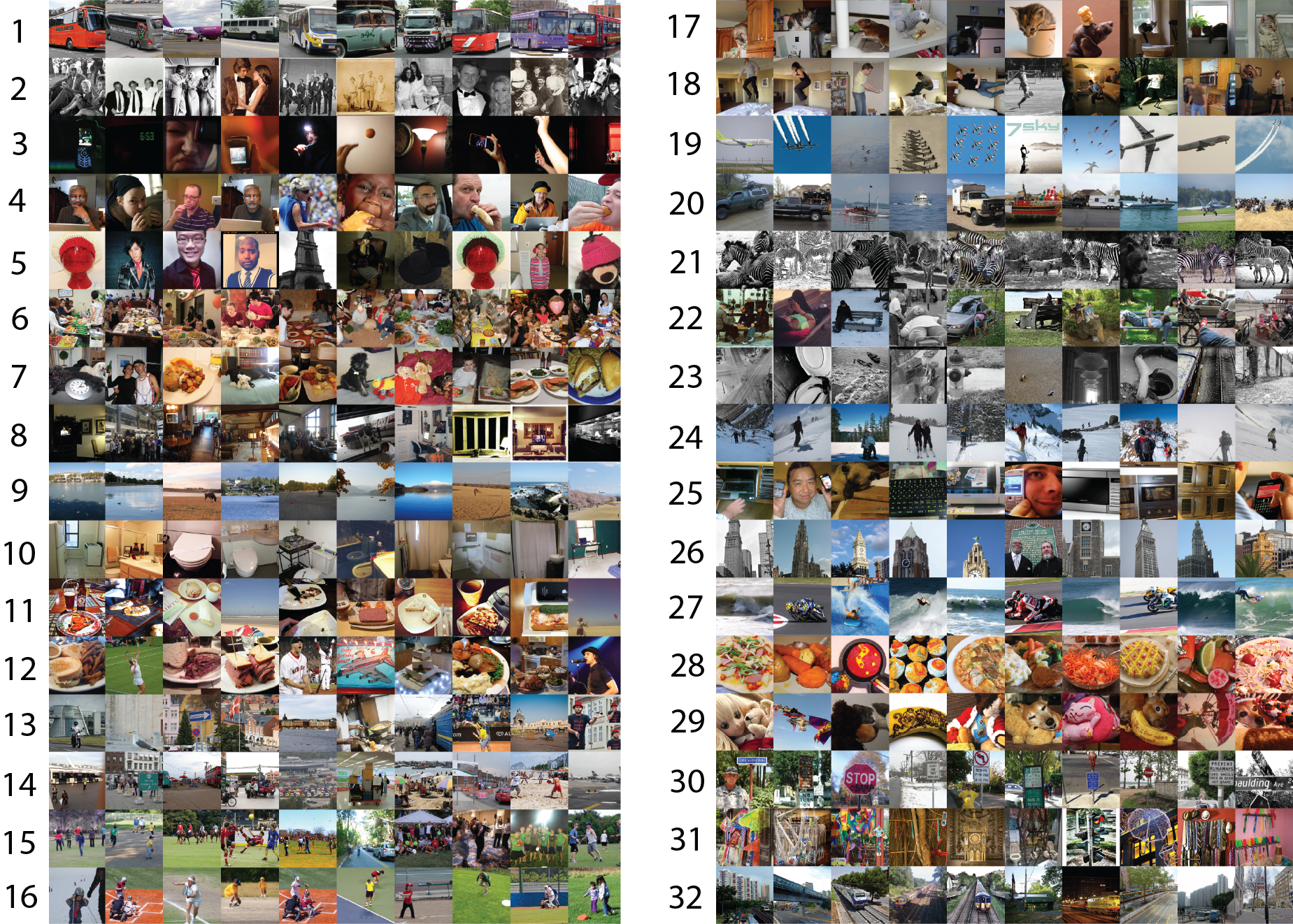}
    \caption{We identified the top-10 nearest neighbour images for each of the 32 dimensions in our shared decodable concepts matrix $W_{SDC}$. Concretely, our fMRI-derived approximations to CLIP space finds the dimensions in CLIP to which the brain preferentially responds in a manner that is shared across all participants. Each dimension displays a level of semantic consistency throughout the associated images, from highly interpretable dimensions of e.g. buildings (dim-26), signs (dim-30), or sports (dim-16) to other dimensions that are superficially consistent yet are harder to label concretely (dim-31). Some dimensions are clearly of a complex semantic nature, e.g. dim-6, which prefers groups of people in an indoor food context. We highlight some of these dimensions further in the main text.}
    \label{top_10_imgs}
\end{figure}

\begin{figure}
    \centering
    \includegraphics[width=14cm]{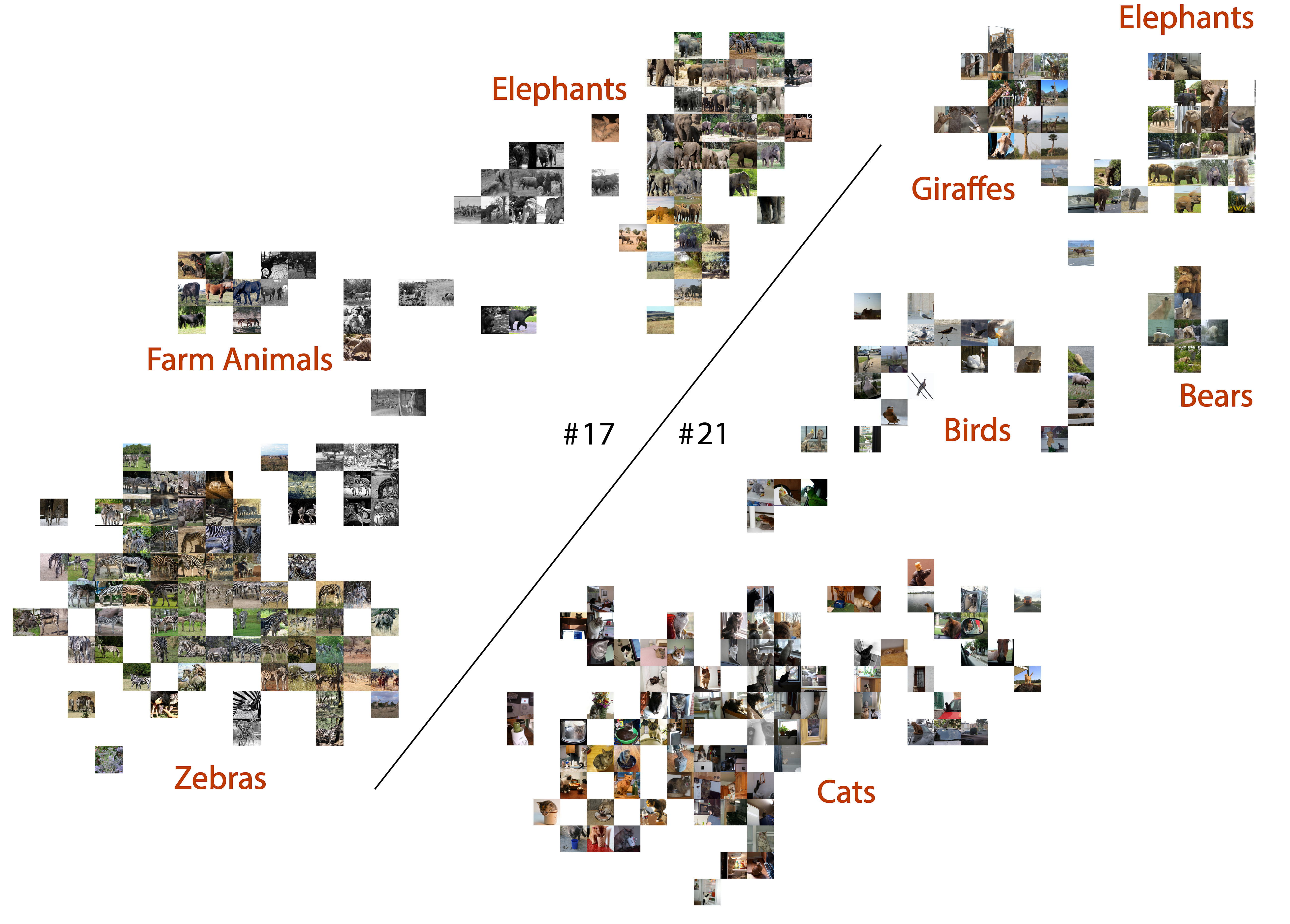}
    \caption{We found two dimensions relating to animals (dim-17: left and dim-21: right) that each contained semantically coherent sub-clusters. We can infer sensitivity to fine-grained shared decodable concept sub-groupings in our method, here between large animals in the wild and smaller animals in different contexts, e.g. cats in a home or birds in the sky.}
    \label{fig:tsne_animals}
\end{figure}

Results for computing top-k accuracy in SDC space appear in Figure~\ref{fig:spaces} (orange bars).  Decoding performance in SDC space is higher than in the original CLIP space, and grows with increasing dimension.  This implies that there are visual concepts available in CLIP space that are decodable from fMRI, but that CLIP contains information not available from the fMRI in our experimental setting. Figure \ref{top_10_imgs} depicts the top 10 nearest-neighbour images for each learned concept vector, namely, each row in $W_{SDC}$ with $c = 32$. We then visualized a selection of concept vectors by projecting a larger selection of nearest-neighbours ($N=250$) down to 2-dimensional space using t-distributed stochastic neighbour embedding (t-SNE) projection. Figure \ref{fig:tsne_animals} outlines two dimensions we found to represent animal concepts. Further dimensions associated with other semantically-coherent concepts are given in the Appendix: food (Fig. \ref{fig:tsne_food}), household rooms (Fig. \ref{fig:tsne_toilets}), buildings (Fig. \ref{fig:tsne_buildings}) and images associated with strong uniform backgrounds (skies, snow-covered mountains etc.) (Fig. \ref{fig:tsne_skies}). The next sections explore this new SDC space for consistency and specificity across participants. 

\section{Brain-Decodable Concepts that Consistently Correspond to Specific Brain areas}\label{brain-decodable-concepts}

We localize each concept dimension from our learned $W_{SDC}$ matrix to a small subset of voxels with a masking procedure that selects sparse voxel sub-groups. We investigate the spatial contiguity and sparsity of concept voxel sub-groups, and the cross-participant and cross-concept consistency via participant- and concept-specific voxel masks, $m_{i,s}$. To test \textit{consistency} across participants, we define a new metric which takes into account the fractional overlap of voxels present in concept masks across the ROIs calculated using the Human Connectome Project Atlas (HCP-MMP1) for each participant. This allows us to determine broad spatial similarity patterns across participants in their native brain spaces (not aligned to an average brain template), allowing smaller voxel subgroups tuned to finer-grained semantic distinctions to be compared, while maintaining idiosyncratic participant-specific functional anatomy.

\subsection{Finding Concept Brain Areas}
For each CLIP concept vector $\bm{w}_i \in \mathbb{R}^{512}$ from $\bm{W}_{\text{THINGS}}$ and $\bm{W}_{\text{SDC}}$, our objective is to find a sparse binary voxel mask $\bm{m}_{i,s} \in \{0, 1\}^v$ ($v$ is the number of voxels) that defines a set of voxels that support the decodability of the concept $\bm{w}_i$ for participant $s \in \{1, \ldots, 8\}$. The masks are derived by fitting a LASSO regression model 

\begin{align}  
\min_{\bm{m}_{i,s}^\text{lasso}} \quad
\frac{1}{2n}
\lVert \bm{X}^\text{Val} \bm{m}^\text{lasso}_{i,s} - 
\bm{Y}^\text{Val}_\text{CLIP} \bm{w}_i \rVert_2^2 + 
\alpha \lVert \bm{m}_{i,s}^\text{lasso} \rVert_1
\end{align}

\begin{figure}
    \centering
    \includegraphics[width=13cm]{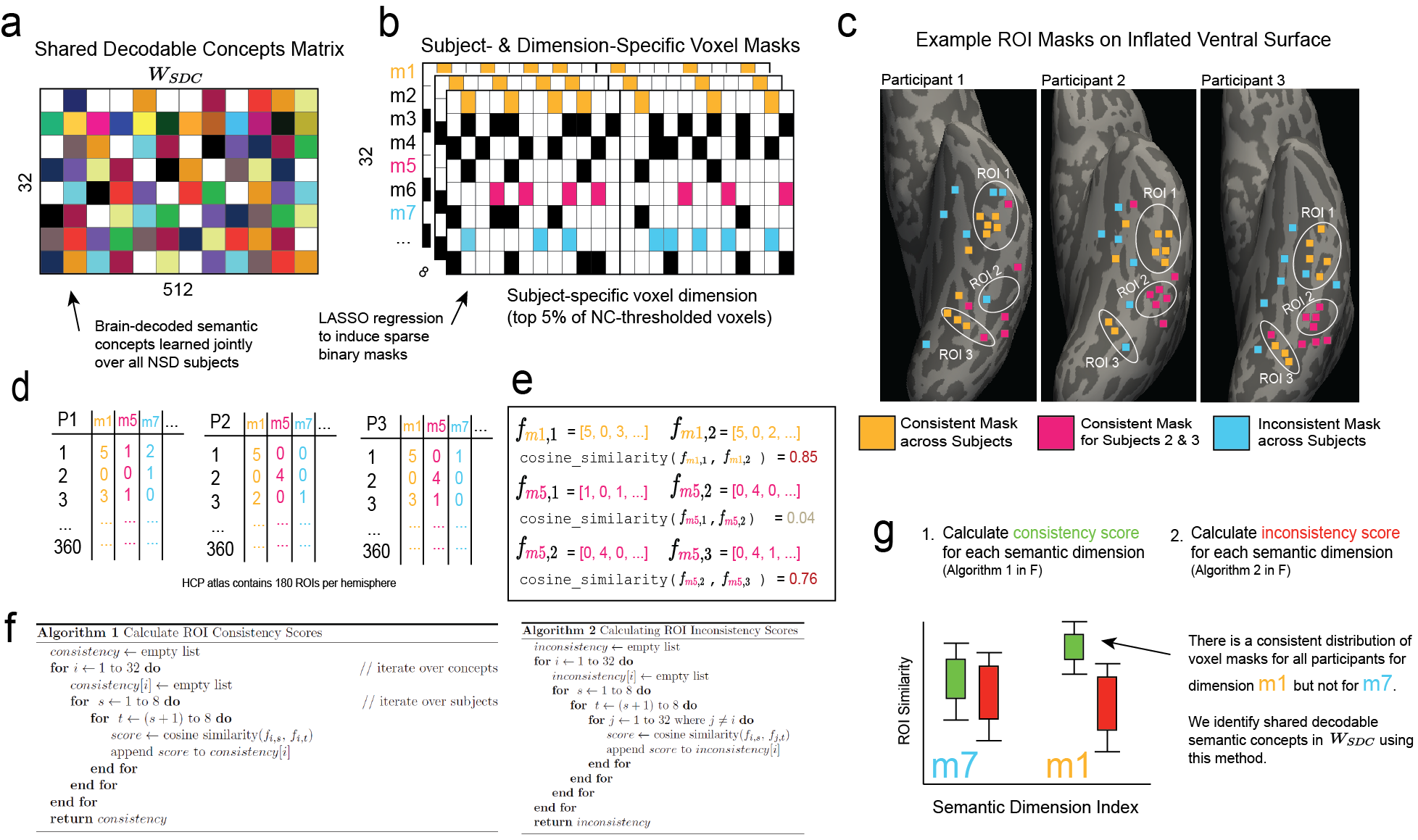}
    \caption{A visual depiction of the algorithmic steps we use in order to calculate ROI-similarity. This method ultimately allows us to identify shared decodable concepts that are consistent across participants and also allow for comparison in participant-specific space due to linking participant-specific masks with their own ROI parcellation, thereby circumventing the need to warp to an average space. This method can therefore be applied to study semantic networks in a participant's native brain space as well as in a cohort of participants. (A) is the shared decodable matrix, $W_{SDC}$. (B) depicts the per-participant voxel masks for each dimension of $W_{SDC}$. (C) An example voxel distribution for 3 concepts of $W_{SDC}$, showing various possible distributions (fully consistent, partially consistent and inconsistent). (D) Each voxel mask's ROI intersection with the HCP atlas is calculated. (E) Cosine similarity of the vectors specified in (D) reveal a consistency metric. (F) Algorithmic steps to calculate within-concept participant consistency and inconsistency. (G) Visual analysis of these values presents a way to detect voxel masks that are more consistent across participants.}
    \label{roi_similarity}
\end{figure}

Here $n$ is the number of data points and the regularization hyperparameter $\alpha = 1e-3$ is used for all concept masks. All non-zero values in $\bm{m}_{i,s}^\text{lasso}$ are set to $1$ to derive the binary mask $\bm{m}_{i,s}$ that is used for further analysis in this section.

\subsection{Evaluation of SDC concept masks}
\begin{figure}
    \centering
    \includegraphics[width=\textwidth]{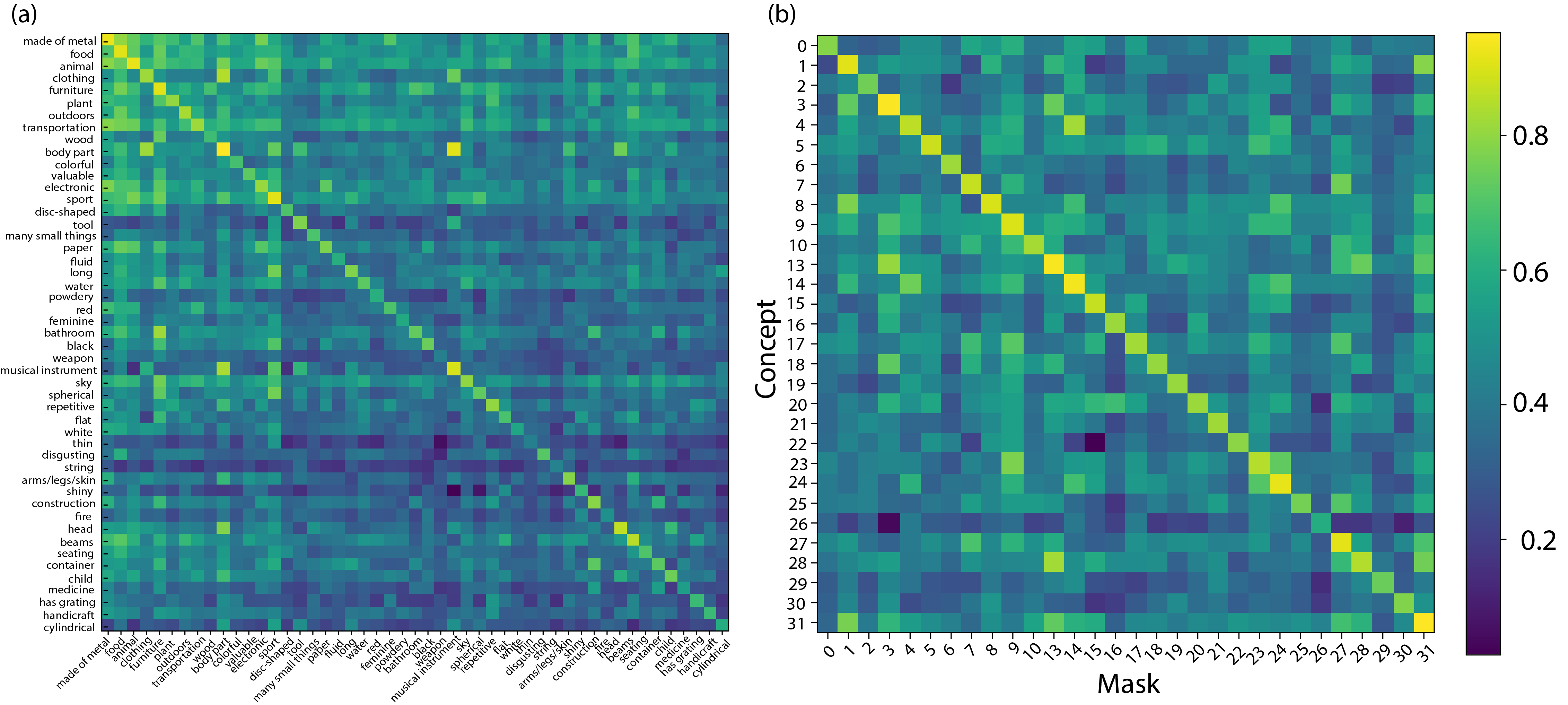}
    \caption{Visualization of the matrices $\bm{D}$ representing the relative change in brain decodability after applying concept masks (defined in Equation \ref{eq:5}). The matrices were constructed separately for $\bm{W}_\text{THINGS}$ (a) and $\bm{W}_\text{SDC}$ (b). We draw attention to the relatively high diagonals of these matrices. This implies that the brain regions represented by masks $\bm{m}_{i,s}$ support the decodability of their corresponding concept vectors $\bm{w}_i$ more than other concepts $\bm{w}_j$ where $j \ne i$.}
    \label{fig:pearsonr_retained}

\end{figure}

\paragraph{Mask Concept Specificity} An initial question is whether the brain regions represented by $\bm{m}_{i,s}$ are truly specific to their corresponding concept vectors $\bm{w}_i$, or do they equally support the decodability of other concepts $\bm{w}_j$ where $j \ne i$? To help answer this, we measure the relative change in brain decodability for a concept $\bm{w}_j$ after applying a mask $\bm{m}_{i,s}$. The matrix of relative changes in brain decodability $\bm{D}^{s}$ for a participant $s$ is constructed as follows:

\begin{align} \label{eq:5}
D_{i, j}^{s} =
\frac
{
    \text{pearsonr}(
        \bm{Y}^\text{Test}_\text{CLIP} \cdot \bm{w}_i, 
        \bm{g}(\bm{X^\text{Test}} \odot \bm{m}_{j, s}) \cdot \bm{w}_i
    )
}
{
    \text{pearsonr}(
        \bm{Y}^\text{Test}_\text{CLIP} \cdot \bm{w}_i, 
        \bm{g}(\bm{X^\text{Test}}) \cdot \bm{w}_i
    )
}
\end{align}
where $\odot$ denotes element-wise multiplication and $\text{pearsonr}(.)$ is the Pearson correlation coefficient. Then the participant-specific $\bm{D}^{s}$ are averaged across participants to obtain $\bm{D}$ which is displayed in Figure \ref{fig:pearsonr_retained}.

\paragraph{Cross Participant Consistency} Next, we investigate whether the masks $\bm{m}_{i, s}$ exhibit consistency in their spatial locations within the brain across participants. The highly sparse and disjoint nature of the masks motivated the use of an ROI-based similarity measure. The \textit{HCP\_MMP1} atlas parcellates the cortical surface into 180 distinct regions per hemisphere. We used this atlas to define ROI fraction vectors $f_{i, s} \in \mathbb{R}^{360}$ for each concept mask $m_{i, s}$. Each element in $f_{i, s}$ represents the number of voxels in $m_{i, s}$ that intersect a particular ROI in the atlas, divided by the total number of voxels in $m_{i, s}$. The similarity of mask regions can be compared across participants by taking the cosine similarity between ROI fraction vectors, i.e. $\text{cosine similarity}(f_{i, s}, f_{j, t})$ where $i, j$ index concepts and $s, t$ index participants. Figure \ref{roi_similarity} outlines the method we use to identify shared decodable concepts.

Using our shared decodable concepts matrix $W_{SDC}$, (Fig. \ref{roi_similarity}a), we use LASSO regularization to induce a participant-specific subset of voxels we call a \textit{voxel mask}, such that each individual participant has a sparse voxel mask for each dimension of $W_{SDC}$ (Fig. \ref{roi_similarity}b). The dimensionalities of the masks per-participant and per-mask are given in \ref{dimensionality}. We demonstrate the procedure by selecting 3 synthetic example dimensions for purposes of illustration, $m1$, $m5$ and $m7$ for three participants in the fMRI dataset. In Fig-\ref{roi_similarity}c we show an example of how these voxel masks for 3 participants might be spatially organised along the inflated ventral surface (images generated using Freesurfer's Freeview program \citep{ref:freesurfer}). Dimension $m1$ (orange voxels) is highly consistent across participants, meaning that it frequently appears in ROI-1 and ROI-3 of each participant. Dimension $m5$ (magenta voxels) is consistent between participants 2 and 3, both appearing in ROI-2 but not in ROI-2 in Participant 1. This represents partial consistency. Finally, dimension $m7$ (blue voxels) represents a situation where the spatial organization of the mask does not intersect any defined ROI consistently across participants.  By taking the 360 automatically-generated ROIs (180 from each hemisphere) from the Human Connectome Project (HCP) Atlas \citep{HCP}, calculated by Freesurfer's \textit{recon-all} program, we count the number of voxels in each mask that intersect with each ROI and for each participant's set of masks. This yields a 360-dimensional vector for each participant and each mask (Figure \ref{roi_similarity}d). We then calculate the cosine similarity between these vectors in order to determine if there is a consistent shared representation of mask voxels across the ROIs as shown in Fig. \ref{roi_similarity}e. For each dimension of $W_{SDC}$, we calculate a list of consistent (high cosine similarity) and inconsistent (low cosine similarity) scores according to the algorithms given in Fig. \ref{roi_similarity}f. 

\begin{figure}
    \begin{subfigure}[b]{.74\textwidth}
        \centering
        \includegraphics[width=13.5cm]{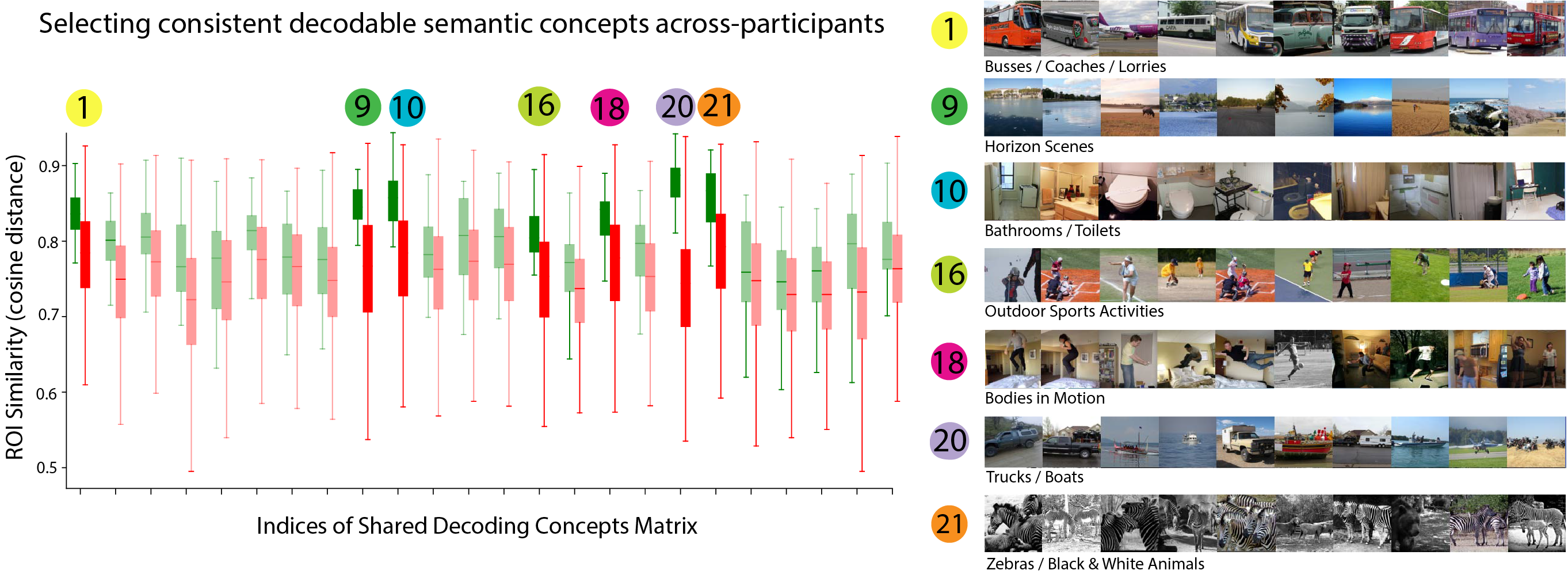}
    \end{subfigure}
    \caption{Identification of concepts from the 32-dimensional SDC space that exhibit higher cross-participant consistency. The green box plots show the distribution of cosine similarities between ROI fraction vectors for a particular concept $i$ across participants, i.e. $\text{Green}_i = \{\text{cossim}(f_{i, s}, f_{i, t}) | s \ne t\}$. The red box plots show the distribution for the same concept $i$ compared to all other concepts for all other participants, i.e $\text{Red}_i = \{\text{cossim}(f_{i, s}, f_{j, t}) | s \ne t, i \ne j\}$. We note that the distributions of the within-concept cross-participant similarities (green) are slightly higher than cross-concept cross-participant similarities (red) for all concepts. Additionally, we draw attention to a set of 7 concepts that are especially consistent and display their representative images on the right. The representative images that have the highest CLIP similarity to the corresponding concept vector $\bm{w}_i$ are selected from the NSD stimulus set.}
    \label{fig:consistent_concepts}
\end{figure}

By comparing the distributions of both sets of results, we identify dimensions that share stable mask distributions across participants (Fig. \ref{roi_similarity}g). For some dimensions of our optimized matrix, we find greatly increased consistency across participants. We selected the top 7 dimensions and plot the 10 most associated images for that dimension in Figure \ref{fig:consistent_concepts}. These shared decodable concept dimensions have associated images that are superficially very distinct but have a clearly consistent semantic interpretation.

\paragraph{Voxel subselection}

The mask optimization procedure does not explicitly encourage the discovery of contiguous voxel subgroups, yet this is exactly what we find when generating masks for each of the decodable semantic concepts. 
This allows us to explore mask distributions across ROIs associated with specific semantically congruent stimuli, such as dimensions often used in fMRI localizer scans. We consistently find that semantic dimensions in our shared decoding space that contain prominent faces, bodies and places have contiguous mask locations in the expected ROIs. These voxel subgroups in our results occupy much smaller and fine-grained areas of ROIs found in localizer experiments, which may be related to processing the various fine-grained semantic distinctions. We expand on this analysis in Appendix \ref{appendix:consistency}. Using the dimensions we found to be most consistent across participants (Figure \ref{fig:consistent_concepts}: right), we further explore the voxel mask locations to identify which brain areas support the semantic dimensions of $W_{SDC}$ that are consistent across participants.

\subsection{Identifying Brain Regions Underpinning Fine-Grained Shared Decodable Concepts}\label{sec:identifying_shared}
\begin{figure}
    \centering
    \includegraphics[width=14cm]{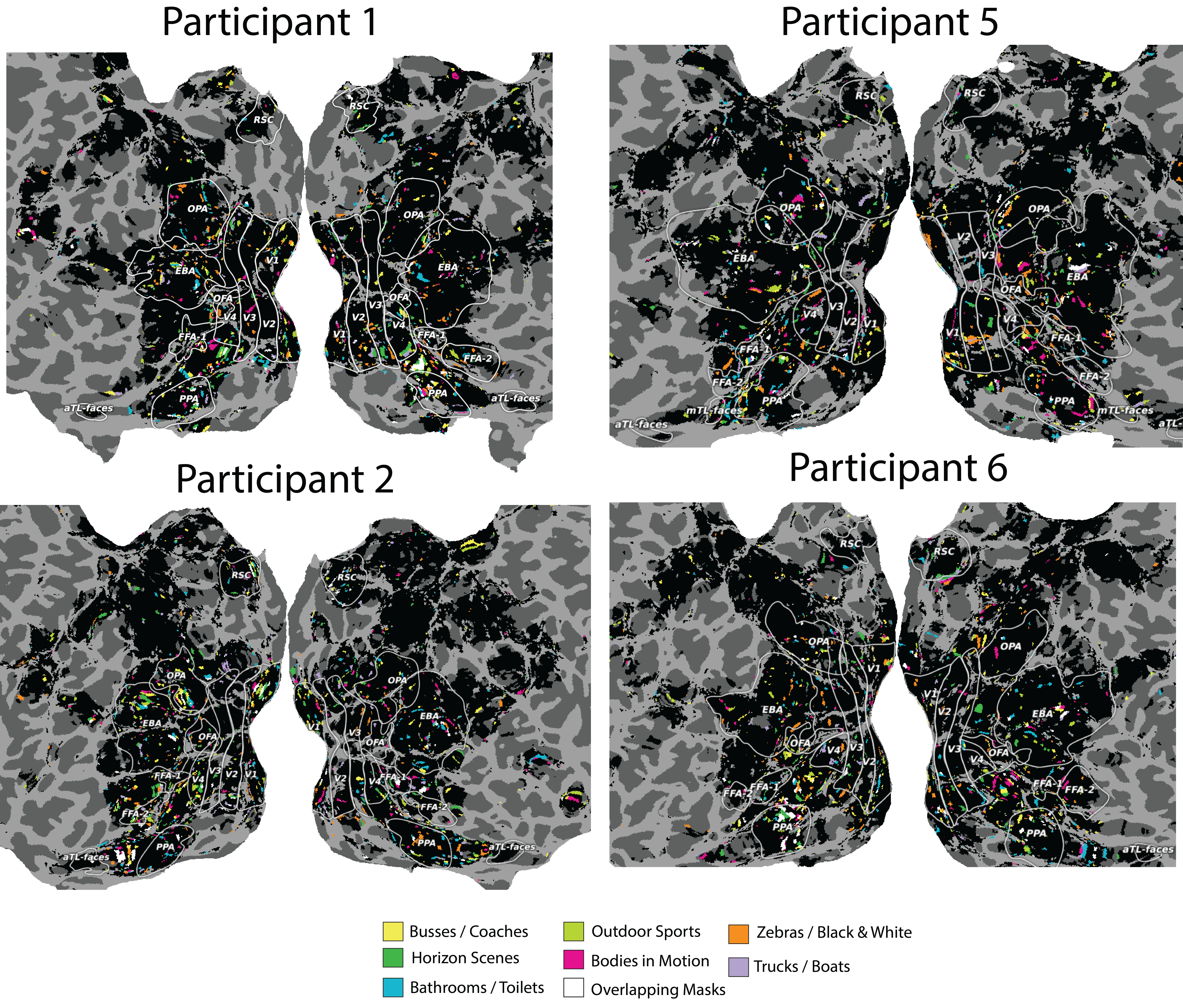}
    \caption{A selection of 4 participants from the Natural Scenes Dataset. For the 7 shared decodable concepts we identified in Figure \ref{fig:consistent_concepts}, we show the color-coded semantic voxel mask locations. Black sections of flat maps represent voxels that passed the 5\% noise ceiling threshold on the training set (see Section \ref{sec:data_prep}). White areas represent overlap of voxel mask locations for multiple concepts.}
    \label{fig:composite2}
\end{figure}

We visualized the spatial locations of voxel masks for key semantic dimensions which we found to be consistent across subjects in Figures \ref{fig:consistent_concepts} and \ref{fig:pearsonr_retained}. We noted several consistencies and plot color-coded voxel masks for a sample of 4 NSD participants in Figure \ref{fig:composite2}. For Participant 1, we found an area where the masks for dimensions 16 and 18 largely overlap. The top images in dimension 16 depict people performing a variety of sports activities, and dimension 18 mainly depicts people bodily motion (i.e. jumping). The brain area where the two corresponding maps overlap is PFcm, which has been associated with the action observation network~\citep{Caspers2010,Urgen2021}. We highlight this region later in Figure \ref{fig:qual_motion}. In the same participant, we also note that masks for dimensions 9 and 20 overlap substantially. The top images in those dimensions contain horizon scenes and trucks/boats, respectively, and are conceptually linked by their outdoor settings. The brain area where these maps overlap is area TF of parahippocampal cortex: importantly, the parahippocampal place area (PPA) is known to represent places and identification of such voxel masks points to specialized subsections that could underlie more fine-grained representations relating to the broad semantic concept of \textit{place}.

The t-SNE results in Section \ref{finding_SDCs} (and Appendix \ref{tsne}) show remarkably coherent semantic networks containing hierarchical representations that cluster together in human-interpretable ways. We further explore the spatial organization of some of these dimensions by inspecting the participant-specific voxel masks learned for these dimensions in order to derive hypotheses for future work out of our data-driven approach. 

\begin{figure}
    \centering
    \includegraphics[width=14cm]{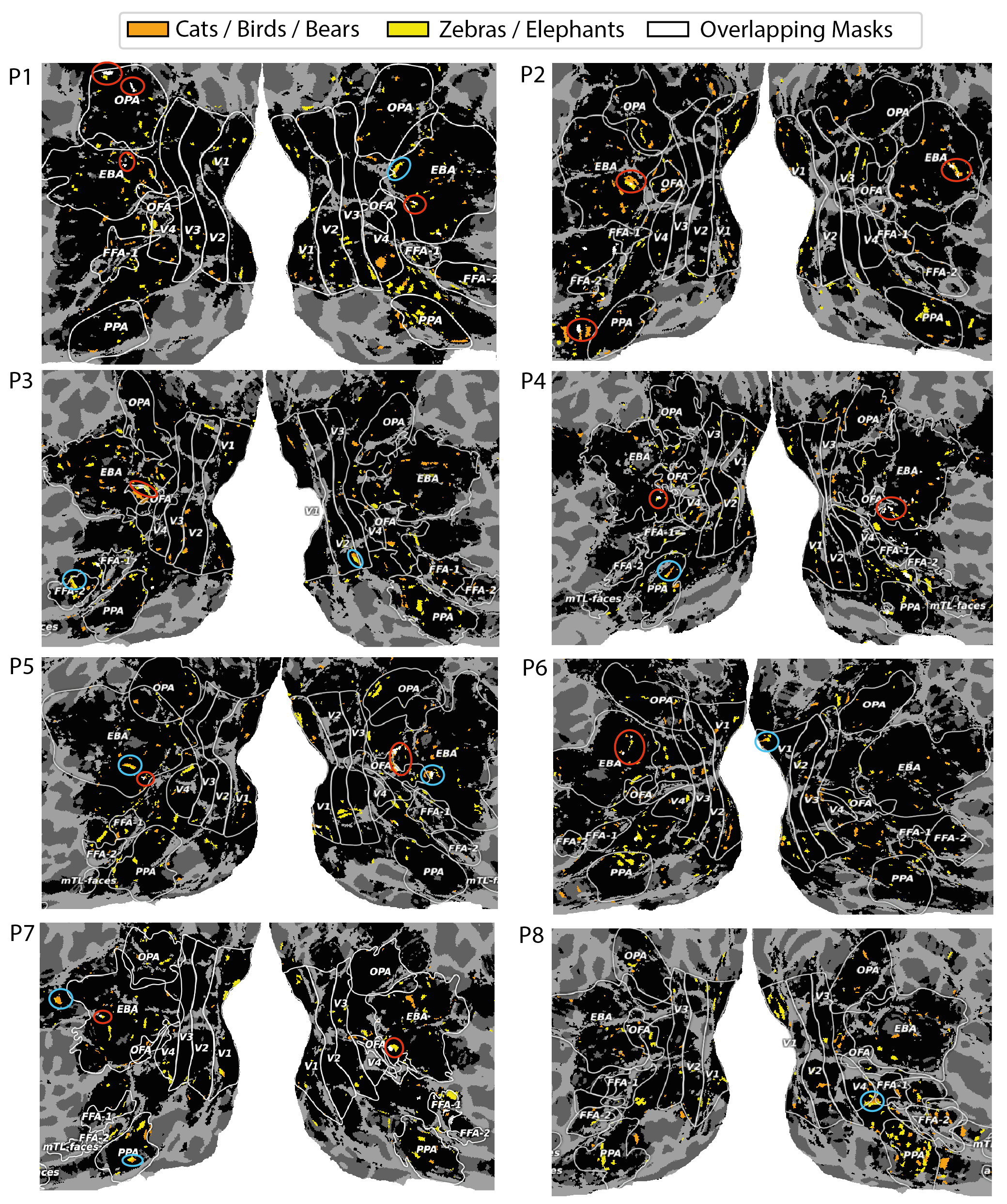}
    \caption{Flat maps of the 8 NSD participants and the voxel masks corresponding to dimensions 17 and 21, the animal dimensions previously referenced in t-SNE plots. Red circles represent areas of overlapping masks (white), typically found in bilateral EBA regions. Blue circles represent adjacent borders of voxels that are non-overlapping but tightly connected. We posit that these voxel masks could be used to explore semantic features shared by both dimensions (animals), as well as those that separate them (animal size, familiarity, typical environments in which they are found).}
    \label{fig:qual_animals}
\end{figure}

Figure \ref{fig:tsne_animals} showed two dimensions related to animals, in which one (dimension 17) contained clusters of cats, birds, bears, giraffes and elephants, while the other (dimension 21) contained clusters of zebras, farm animals and (again) elephants. Figure \ref{fig:qual_animals} shows the 8 participants' voxel masks for these two dimensions. We first note that the masks are largely non-overlapping, but we do see bilateral sections of the EBA selected (red circles) consistently across participants, though often at different extrema of the EBA's boundaries, reflecting an individual's functional / anatomical individuality. Furthermore, for participants 1,2,3,5,6,8 we see much greater voxel selection in PPA for dimension 21 (yellow), which is the dimension that represents animals in the wild, compared to dimension 17, which is largely related to indoor cats or birds in the sky. We also find in some participants that voxel masks are learned that share adjoining yet non-overlapping continua (blue circles). We believe this an interesting finding, after discovering the most selective images associated with these components were semantically linked to \textit{animals}. We perform the same analysis on the \textit{food} dimension that we identified (dimension 28) in Appendix \ref{app:food_flatmaps}. 

\subsubsection{Identifying Within-Participant Brain Regions Underpinning Shared Semantics}

\begin{figure}
    \centering
    \includegraphics[width=14cm]{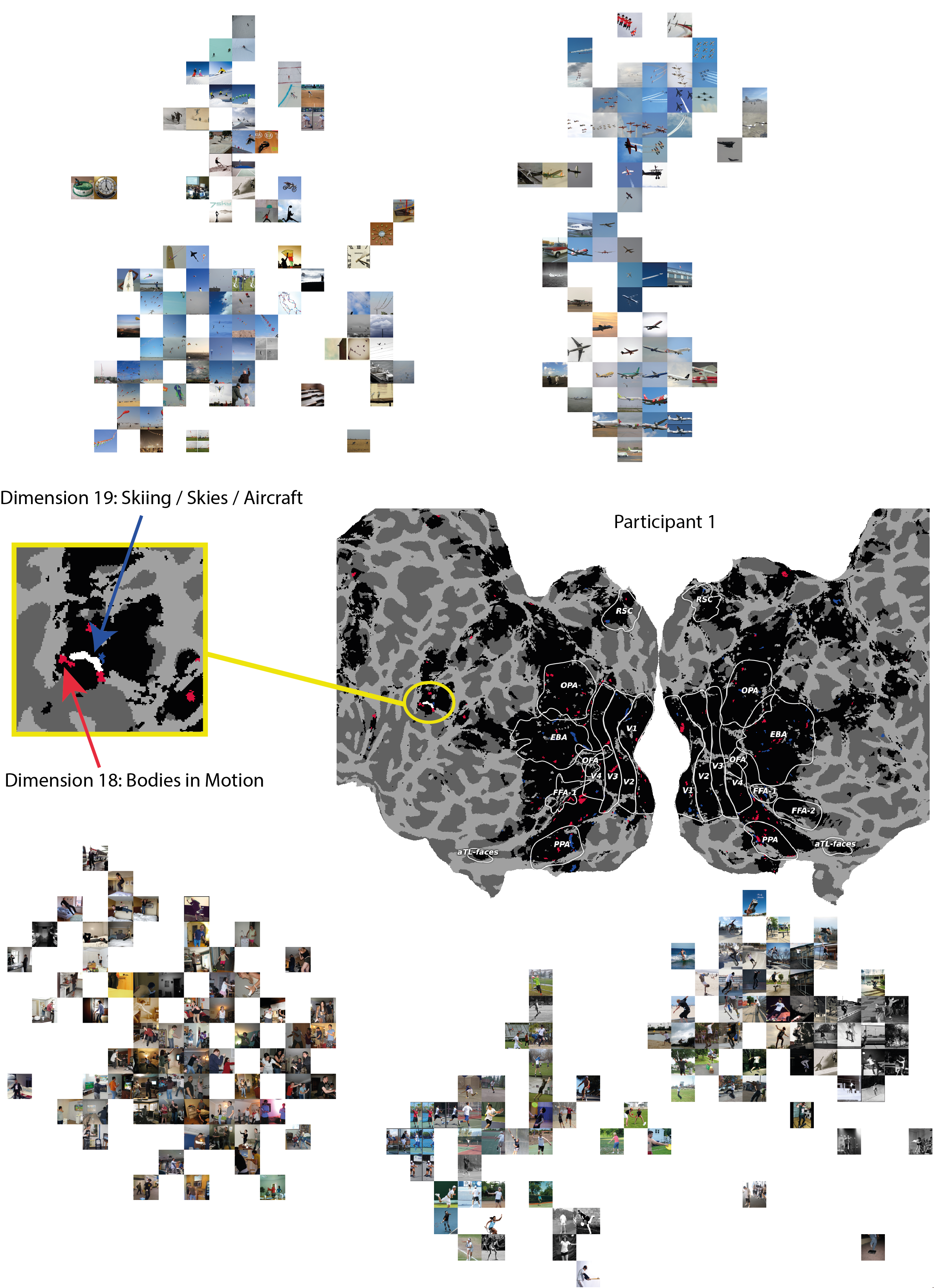}
    \caption{Top: t-SNE clusters related to dimension 19 (skiing / skies / aircraft). Middle: Participant 1's flat map, highlighting area PFcm in the left hemisphere, contains overlapping voxels relating to the dimension-specific learned voxel masks for this participant. Bottom: t-SNE clusters related to dimension 18. Both dimensions encode action observation, a feature previously reported for that same patch of cortex (PFcm).}
    \label{fig:qual_motion}
\end{figure}

In Section \ref{sec:identifying_shared} it was suggested that Participant 1's voxel mask distribution to dimension 10 ("bodies in motion") had found a large continuous patch of voxels in region PFcm, a region that had previously been linked to bilateral activation when processing action / activity \citep{Urgen2021}. This voxel mask largely overlapped with the voxel mask for dimension 16 ("outdoor sports"). We looked for other semantic concepts that implied movement / action and selected dimension 19, due to the presence of skiers, snowboarders and skateboarders in the t-SNE visualisation of the top images in that cluster (see \ref{app:skies}). Encouragingly, we find that this semantic dimension has also induced a learned voxel mask that overlaps with the same patch of cortex in this area of interest. This appears to serve as a signature of observed action when processing images, given the varied surface level features of the visual images in each dimension. Figure \ref{fig:qual_motion} shows the flat map representation of Participant 1's cortex (middle), with the voxel mask locations for these two dimensions overlaid. We further examined the t-SNE clusters for these dimensions (top: dimension 19, bottom: dimension 18). When only plotting the top-10 nearest neighbours earlier in Figure \ref{top_10_imgs}, we only identified the concepts implying body motion indoors, but extending this to a larger numbers of images, we also see that this dimension captures bodies in motion during sports, beyond just jumping. We note that virtually all images connected to dimension 18 involve a form of action, while a portion of dimension 19 also overlap in the same semantic regions (skiing, snowboarding, skateboarding, kite-flying). We find overlap in the left PFcm region, which has been previously associated with action observation (yellow box). The patch of cortex where both dimension masks overlap reveal that the mask for dimension 19 is smaller than that of dimension 18, which makes sense as only a subset of the images associated with dimension 19 are specific to action observation.

These t-SNE clusters reveal two sets of images that are superficially different but we find semantic overlaps that can be linked to adjoining / overlapping patches of cortex that sparse voxel masks can detect, which can then be related to prior literature that previously found evidence that the same region supports the same semantic overlap we observed in the image dimensions, namely action observation. We find that for a few participants that the contrast of these two dimensions results in overlapping continuous strips in PPA (not shown). We envisage that the method we present to identify shared decodable concepts can also be used to map out within-participant fine-grained semantic networks and we demonstrate an example of this on our initial results here.

\section{Discussion}

We introduced a new data-driven method that uses the CLIP language-image foundation model~\citep{CLIP} to identify sets of voxels in the human brain that are specialized for representing different concepts in images. Using this method, we identified several putative new concept-encoding networks that were consistent between individuals. Our method also identified concept-encoding regions that were previously known (e.g., regions specialized for faces, places, and food), which increased our confidence in the new regions uncovered by our method.

Importantly, our new method enables us to compare fine-grained participant-specific concept-encoding networks (defined as sets of voxels) without needing to apply any compression to the brain imaging data. These networks can then be mapped to 180 different anatomically-defined brain regions in each hemisphere, enabling us to relate our concept-encoding networks to known brain anatomy. At the same time, by virtue of its fine-grained resolution, our method can identify concept-encoding networks that span multiple anatomical brain regions, and those that occupy portions of known brain regions.

Among the new concepts-encoding networks we identified are networks specialized for jumping bodies, transport, scenes with strong horizons, room internal (toilet/bathroom). These are clearly important concepts, but their representations are ones that would not be readily identified using standard methods that attempt to tie brain areas to semantic concepts. This is because the space of potential concepts to explore is so large that a hypothesis-driven ``guess and check'' approach is likely to miss many important concepts. For this reason, our data-driven approach represents an important advance, and highlights the utility of modern AI systems (e.g., the CLIP language-image foundation model) for advancing cognitive neuroscience. Importantly, the method we introduced is quite general: while we have applied it here to CLIP embeddings and fMRI data, future work could apply the same method to embeddings generated by other AI systems (not just CLIP), and/or to other brain recording modalities (not just fMRI).

All neural recordings are from a previously released public dataset~\citep{NSD}.  The recordings were collected with the oversight of the institutional review boards at the institute where the experiments were performed. Our work has the potential to enable new methods for decoding human brain activity using non-invasive methods. Such methods could have substantial impacts on society. On the positive side, these methods could assist in diagnosing psychiatric disorders, or in helping individuals with locked-in syndrome or related disorders to better communicate. On the other hand, the same brain decoding methods could pose privacy concerns. As with all emerging technologies, responsible deployment is needed in order for society as a whole to obtain maximum benefit while mitigating risk.

There are several limitations of this work, and decoding experiments in general.  Firstly, we use fMRI, and so can only recover the shared decodable concepts available in that fMRI space.  That is, if concepts are not encoded by the BOLD signal as measurable by fMRI, we will not be able to recover them.  On the other hand, the methods we develop here could be applied to other neural recording modalities that more directly measure neural activity, including electrocorticography. Consequently, this first limitation is one that could be overcome in future studies.

Secondly, we are also only able to uncover concepts that appear in the CLIP space spanned by the images in the stimulus set. This means that, while our method is more flexible than the hypothesis-driven ones used by previous studies~\citep{Kanwisher1997}, some important concept representations could still be missed by our method. Future work could address this limitation by using even larger stimulus sets. Finally, we report on a few probable localizations of concepts. These provide clear hypotheses for future experiments that should be targeted at attempting to confirm these findings. This emphasizes that our new data-driven method is not a replacement for the hypothesis-driven approach to identifying concept-encoding regions in the human brain: rather, it serves as a data-driven hypothesis generator that can accelerate the task of understanding how our brains represent important information about the world around us.

\newpage
\bibliography{refs.bib}  

\begin{thebibliography}{24}
\providecommand{\natexlab}[1]{#1}
\providecommand{\url}[1]{\texttt{#1}}
\expandafter\ifx\csname urlstyle\endcsname\relax
  \providecommand{\doi}[1]{doi: #1}\else
  \providecommand{\doi}{doi: \begingroup \urlstyle{rm}\Url}\fi

\bibitem[Allen et~al.(2022)Allen, St-Yves, Wu, Breedlove, Prince, Dowdle, Nau,
  Caron, Pestilli, Charest, Hutchinson, Naselaris, and Kay]{NSD}
E.~J. Allen, G.~St-Yves, Y.~Wu, J.~L. Breedlove, J.~S. Prince, L.~T. Dowdle,
  M.~Nau, B.~Caron, F.~Pestilli, I.~Charest, J.~B. Hutchinson, T.~Naselaris,
  and K.~Kay.
\newblock A massive 7t fmri dataset to bridge cognitive neuroscience and
  artificial intelligence.
\newblock \emph{Nature Neuroscience}, 25\penalty0 (1), 2022.
\newblock \doi{10.1038/s41593-021-00962-x}.

\bibitem[Caspers et~al.(2010)Caspers, Zilles, Laird, and Eickhoff]{Caspers2010}
S.~Caspers, K.~Zilles, A.~R. Laird, and S.~B. Eickhoff.
\newblock Ale meta-analysis of action observation and imitation in the human
  brain.
\newblock \emph{NeuroImage}, 50\penalty0 (3):\penalty0 1148--1167, 2010.
\newblock ISSN 1053-8119.
\newblock \doi{https://doi.org/10.1016/j.neuroimage.2009.12.112}.
\newblock URL
  \url{https://www.sciencedirect.com/science/article/pii/S1053811909014013}.

\bibitem[Epstein and Kanwisher(1998)]{Epstein1998}
R.~Epstein and N.~Kanwisher.
\newblock A cortical representation of the local visual environment.
\newblock \emph{Nature}, 392\penalty0 (6676):\penalty0 598--601, 1998.

\bibitem[Fischl(2012)]{ref:freesurfer}
B.~Fischl.
\newblock Freesurfer.
\newblock \emph{NeuroImage}, 62\penalty0 (2):\penalty0 774--781, 2012.
\newblock ISSN 1053-8119.
\newblock 20 YEARS OF fMRI.

\bibitem[Gao et~al.(2015)Gao, Huth, Lescroart, and Gallant]{pycortex}
J.~S. Gao, A.~G. Huth, M.~D. Lescroart, and J.~L. Gallant.
\newblock Pycortex: an interactive surface visualizer for fmri.
\newblock \emph{Frontiers in Neuroinformatics}, 9, 2015.
\newblock ISSN 1662-5196.
\newblock \doi{10.3389/fninf.2015.00023}.

\bibitem[Glasser et~al.(2016)Glasser, Coalson, Robinson, Hacker, Harwell,
  Yacoub, Ugurbil, Andersson, Beckmann, Jenkinson, Smith, and Essen]{HCP}
M.~Glasser, T.~Coalson, E.~Robinson, C.~Hacker, J.~Harwell, E.~Yacoub,
  K.~Ugurbil, J.~Andersson, C.~Beckmann, M.~Jenkinson, S.~Smith, and D.~v.
  Essen.
\newblock A multi-modal parcellation of human cerebral cortex.
\newblock \emph{Nature (London)}, 536\penalty0 (7615):\penalty0 171--8, 2016.
\newblock ISSN 0028-0836.

\bibitem[Hamilton and Huth(2020)]{Hamilton2020}
L.~S. Hamilton and A.~G. Huth.
\newblock The revolution will not be controlled: natural stimuli in speech
  neuroscience.
\newblock \emph{Language, Cognition and Neuroscience}, 35\penalty0
  (5):\penalty0 573--582, 2020.
\newblock \doi{10.1080/23273798.2018.1499946}.
\newblock PMID: 32656294.

\bibitem[Hanson and Halchenko(2008)]{Hanson2008}
S.~J. Hanson and Y.~O. Halchenko.
\newblock Brain reading using full brain support vector machines for object
  recognition: there is no {``}face{''} identification area.
\newblock \emph{Neural Computation}, 20:\penalty0 486--503, 2008.
\newblock ISSN 0899-7667.
\newblock \doi{10.1162/neco.2007.09-06-340}.

\bibitem[Haxby et~al.(2001)Haxby, Gobbini, Furey, Ishai, Schouten, and
  Pietrini]{haxby_2001}
J.~V. Haxby, M.~I. Gobbini, M.~L. Furey, A.~Ishai, J.~L. Schouten, and
  P.~Pietrini.
\newblock Distributed and overlapping representations of faces and objects in
  ventral temporal cortex.
\newblock \emph{Science}, 293\penalty0 (5539):\penalty0 2425--2430, 2001.

\bibitem[Hebart et~al.(2023)Hebart, Contier, Teichmann, Rockter, Zheng, Kidder,
  Corriveau, Vaziri-Pashkam, and Baker]{THINGS}
M.~Hebart, O.~Contier, L.~Teichmann, A.~H. Rockter, C.~Y. Zheng, A.~Kidder,
  A.~Corriveau, M.~Vaziri-Pashkam, and C.~I. Baker.
\newblock Things-data, a multimodal collection of large-scale datasets for
  investigating object representations in human brain and behavior.
\newblock \emph{eLife}, 12, 2023.

\bibitem[Hubel and Wiesel(1959)]{Hubel1959}
D.~H. Hubel and T.~N. Wiesel.
\newblock Receptive fields of single neurons in the cat's striate cortex.
\newblock \emph{Journal of Physiology}, 148:\penalty0 574--591, 1959.

\bibitem[Jain et~al.(2023)Jain, Wang, Henderson, Lin, Prince, Tarr, and
  Wehbe]{jain2023selectivity}
N.~Jain, A.~Wang, M.~M. Henderson, R.~Lin, J.~S. Prince, M.~J. Tarr, and
  L.~Wehbe.
\newblock Selectivity for food in human ventral visual cortex.
\newblock \emph{Communications Biology}, 6\penalty0 (1), 2023.
\newblock \doi{10.1038/s42003-023-04546-2}.

\bibitem[Kanwisher et~al.(1997)Kanwisher, McDermott, and Chun]{Kanwisher1997}
N.~Kanwisher, J.~McDermott, and M.~M. Chun.
\newblock The fusiform face area: A module in human extrastriate cortex
  specialized for face perception.
\newblock \emph{Journal of Neuroscience}, 17\penalty0 (11):\penalty0
  4302--4311, 1997.
\newblock ISSN 0270-6474.
\newblock \doi{10.1523/JNEUROSCI.17-11-04302.1997}.
\newblock URL \url{https://www.jneurosci.org/content/17/11/4302}.

\bibitem[Khosla et~al.(2022)Khosla, Murty, and Kanwisher]{kanwisher2022food}
M.~Khosla, N.~A.~R. Murty, and N.~Kanwisher.
\newblock {{A} highly selective response to food in human visual cortex
  revealed by hypothesis-free voxel decomposition}.
\newblock \emph{Current Biology}, 32\penalty0 (19):\penalty0 4159--4171, 2022.

\bibitem[Kriegeskorte et~al.(2009)Kriegeskorte, Simmons, Bellgowan, and
  Baker]{doubledipping}
N.~Kriegeskorte, W.~Simmons, P.~Bellgowan, and C.~Baker.
\newblock Circular analysis in systems neuroscience: The dangers of double
  dipping.
\newblock \emph{Nature neuroscience}, 12:\penalty0 535--40, 05 2009.

\bibitem[Lin et~al.(2014)Lin, Maire, Belongie, Bourdev, Girshick, Hays, Perona,
  Ramanan, Zitnick, and Dollár]{coco}
T.-Y. Lin, M.~Maire, S.~Belongie, L.~Bourdev, R.~Girshick, J.~Hays, P.~Perona,
  D.~Ramanan, C.~L. Zitnick, and P.~Dollár.
\newblock Microsoft coco: Common objects in context, 2014.
\newblock cite arxiv:1405.0312Comment: 1) updated annotation pipeline
  description and figures; 2) added new section describing datasets splits; 3)
  updated author list.

\bibitem[Matusz et~al.(2019)Matusz, Dikker, Huth, and Perrodin]{Matusz2019}
P.~J. Matusz, S.~Dikker, A.~G. Huth, and C.~Perrodin.
\newblock {Are We Ready for Real-world Neuroscience?}
\newblock \emph{Journal of Cognitive Neuroscience}, 31\penalty0 (3):\penalty0
  327--338, 03 2019.
\newblock ISSN 0898-929X.
\newblock \doi{10.1162/jocn_e_01276}.
\newblock URL \url{https://doi.org/10.1162/jocn\_e\_01276}.

\bibitem[McCandliss et~al.(2003)McCandliss, Cohen, and Dehaene]{McCandliss2003}
B.~D. McCandliss, L.~Cohen, and S.~Dehaene.
\newblock The visual word form area: expertise for reading in the fusiform
  gyrus.
\newblock \emph{Trends in Cognitive Sciences}, 7\penalty0 (7):\penalty0
  293--299, 2003.
\newblock ISSN 1364-6613.
\newblock \doi{https://doi.org/10.1016/S1364-6613(03)00134-7}.
\newblock URL
  \url{https://www.sciencedirect.com/science/article/pii/S1364661303001347}.

\bibitem[Noppeney(2008)]{Uta_tools}
U.~Noppeney.
\newblock The neural systems of tool and action semantics: A perspective from
  functional imaging.
\newblock \emph{Journal of Physiology-Paris}, 102\penalty0 (1):\penalty0
  40--49, 2008.

\bibitem[Pennock et~al.(2023)Pennock, Racey, Allen, Wu, Naselaris, Kay,
  Franklin, and Bosten]{Pennock2023}
I.~M. Pennock, C.~Racey, E.~J. Allen, Y.~Wu, T.~Naselaris, K.~N. Kay,
  A.~Franklin, and J.~M. Bosten.
\newblock Color-biased regions in the ventral visual pathway are food
  selective.
\newblock \emph{Current Biology}, 33\penalty0 (1):\penalty0 134--146, 2023.

\bibitem[Prince et~al.(2022)Prince, Charest, Kurzawski, Pyles, Tarr, and
  Kay]{GLMsingle}
J.~S. Prince, I.~Charest, J.~W. Kurzawski, J.~A. Pyles, M.~J. Tarr, and K.~N.
  Kay.
\newblock Improving the accuracy of single-trial fmri response estimates using
  glmsingle.
\newblock \emph{eLife}, 11:\penalty0 e77599, nov 2022.
\newblock ISSN 2050-084X.
\newblock \doi{10.7554/eLife.77599}.

\bibitem[Radford et~al.(2021)Radford, Kim, Hallacy, Ramesh, Goh, Agarwal,
  Sastry, Askell, Mishkin, Clark, Krueger, and Sutskever]{CLIP}
A.~Radford, J.~W. Kim, C.~Hallacy, A.~Ramesh, G.~Goh, S.~Agarwal, G.~Sastry,
  A.~Askell, P.~Mishkin, J.~Clark, G.~Krueger, and I.~Sutskever.
\newblock Learning transferable visual models from natural language
  supervision.
\newblock \emph{CoRR}, abs/2103.00020, 2021.

\bibitem[Urgen and Orban(2021)]{Urgen2021}
B.~A. Urgen and G.~A. Orban.
\newblock The unique role of parietal cortex in action observation: Functional
  organization for communicative and manipulative actions.
\newblock \emph{NeuroImage}, 237:\penalty0 118220, 2021.
\newblock ISSN 1053-8119.
\newblock \doi{https://doi.org/10.1016/j.neuroimage.2021.118220}.

\bibitem[van~den Oord et~al.(2018)van~den Oord, Li, and Vinyals]{infoNCE}
A.~van~den Oord, Y.~Li, and O.~Vinyals.
\newblock Representation learning with contrastive predictive coding.
\newblock \emph{CoRR}, abs/1807.03748, 2018.

\end{thebibliography}

\appendix

\section{Appendix}

\subsection{Consistency Check with Faces, Bodies and Place Images}\label{appendix:consistency}

\begin{figure}
    \centering
    \includegraphics[width=11cm]{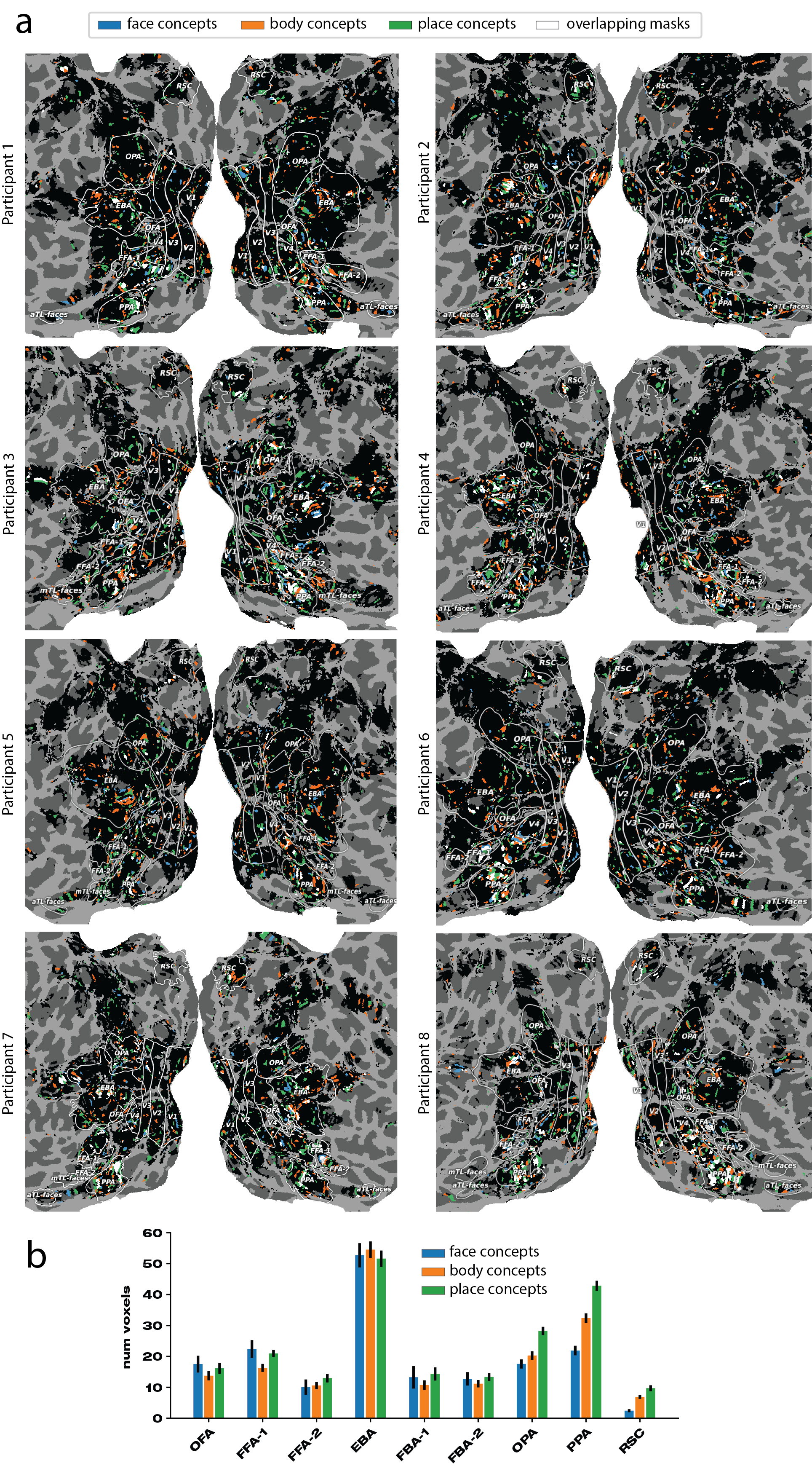}
    \caption{(A) Grouping the mask locations for semantic dimensions that contained consistent face, body and place representations on cortical flat maps. Black areas represent voxels that passed the 5\% noise ceiling threshold. (B) Averaged number of overlapping mask locations with functional localizer regions (provided in NSD). Error bars represent SEM over participants, revealing highly consistent distributions for these concept types.}
    \label{consistency_check}
\end{figure}

In order to verify that our masking procedure captures known functional localization of various high-level visual concepts, such as faces, places, and bodies, we use the functional localizer scans present in NSD for these categories and calculate the overlap with our participant- and dimension-specific voxel masks. To do this, we use our derived $W_{SDC}$ matrix of 32 concepts that are found via fMRI-decoding into CLIP-space, where each dimension represents a potentially shared decodable concept derived from brain responses, revealing the semantic tuning sensitivity to visual concepts that are found to be important. For each dimension of $W_{SDC}$, we extract the top-10 CLIP images that are nearest neighbours to the fMRI-decoded CLIP embeddings. We then identify a set of indices that we expect to activate voxel locations in the participant-specific NSD functional localizer maps. We plot these flat maps for each participant (generated in PyCortex \citep{pycortex}) along with a histogram (average over participants) of voxel mask locations that overlap with areas associated with higher-level functional ventral visual cortex. Figure \ref{top_10_imgs} shows the top-10 associated images with each of the shared decodable concepts. Figure \ref{consistency_check} shows the participant-specific maps and averaged histogram.

From Figure \ref{top_10_imgs}, we associate the three high-level functional categories of interest to the following indices: $face \in \{4, 5\}$, $place \in \{8, 9, 10, 14, 32\}$ and $body \in \{2, 15, 16, 18, 24\}$.

We clearly see that the mask dimensions for the higher-level concepts of faces, bodies and places do largely overlap with areas in which there is an \textit{a priori} expectation of overlap. The images contained in NSD are largely confounded in that each image typically represents not just a single semantic concept. For example, images containing bodies will often contain faces, and most likely vice-versa (but head-shot images would be an exception to this). Furthermore, images of people are often in a place-context, whether inside, outside or in the proximity of an area. We see this fact in our mask distributions, too. In \textit{place} images, such as dimension nine in Figure \ref{top_10_imgs}, where we see strong horizons, there is less confounding with \textit{face} and \textit{body} areas, as place contexts can easily exist without humans, but the reverse is often not true. We see the effect of this in that the place ROIs typically contain the closest association with pooled masks over our identified place dimensions (green bar is higher for OPA, PPA, RSC). These results show that even in a dataset of highly confounded images, the mask locations we derive in our procedure align well with results and expectations from prior literature, serving as a consistency check. Our procedure, however, reveals a vastly more fine-grained set of results across a wider range of cortex, linking (among other regions) functional ROIs together differentially, in order map out a distributed coding of semantically-complex visual inputs that can be used to explore semantic networks between-participants but also across-participants. 

\subsection{Specification of participant-Specific Data Dimensions}\label{dimensionality}

In Section 2.2 we specified that the number of voxels, per-participant, that passed the 5\% noise ceiling thresholds were used as inputs into the initial fMRI decoding algorithm (converting fMRI data to brain-derived CLIP embeddings). The range of voxels that passed this threshold was specified to be in the range of 15-30k. Later on in Section 4.1, we discuss per-participant per-dimension masks (each participant has a specific mask for each of the shared decodable concepts derived in $W_{SDC}$). We specify the dimensionality of voxel inputs and mask dimensions for each participant in Table \ref{tab:voxel_counts}.

\begin{table}[]
\begin{tabular}{|c|c|c|c|c|c|c|c|c|}
\hline
\multicolumn{1}{|l|}{}          & \textbf{P1} & \textbf{P2} & \textbf{P3} & \textbf{P4} & \textbf{P5} & \textbf{P6} & \textbf{P7} & \textbf{P8} \\ \hline
\textbf{Voxels \textgreater NC} & 26,753      & 27,790      & 20,562      & 21,417      & 25,490      & 30,542      & 15,648      & 15,265      \\ \hline
\textbf{Mask 1}                 & 372         & 332         & 352         & 346         & 289         & 459         & 299         & 284         \\ \hline
\textbf{Mask 2}                 & 224         & 243         & 251         & 235         & 194         & 216         & 190         & 176         \\ \hline
\textbf{Mask 3}                 & 256         & 219         & 231         & 246         & 230         & 244         & 176         & 176         \\ \hline
\textbf{Mask 4}                 & 253         & 232         & 254         & 229         & 217         & 268         & 215         & 180         \\ \hline
\textbf{Mask 5}                 & 239         & 199         & 212         & 206         & 187         & 243         & 177         & 154         \\ \hline
\textbf{Mask 6}                 & 322         & 290         & 324         & 284         & 274         & 354         & 278         & 299         \\ \hline
\textbf{Mask 7}                 & 207         & 227         & 178         & 190         & 191         & 196         & 172         & 143         \\ \hline
\textbf{Mask 8}                 & 258         & 254         & 250         & 188         & 219         & 244         & 208         & 175         \\ \hline
\textbf{Mask 9}                 & 292         & 343         & 362         & 305         & 274         & 304         & 274         & 262         \\ \hline
\textbf{Mask 10}                & 342         & 353         & 345         & 323         & 363         & 404         & 327         & 288         \\ \hline
\textbf{Mask 13}                & 295         & 291         & 302         & 275         & 316         & 297         & 244         & 241         \\ \hline
\textbf{Mask 14}                & 272         & 254         & 286         & 261         & 242         & 236         & 223         & 232         \\ \hline
\textbf{Mask 15}                & 307         & 314         & 325         & 257         & 232         & 281         & 224         & 210         \\ \hline
\textbf{Mask 16}                & 255         & 235         & 242         & 233         & 203         & 226         & 222         & 155         \\ \hline
\textbf{Mask 17}                & 291         & 326         & 276         & 261         & 229         & 311         & 209         & 231         \\ \hline
\textbf{Mask 18}                & 245         & 304         & 346         & 280         & 263         & 308         & 282         & 222         \\ \hline
\textbf{Mask 19}                & 204         & 199         & 197         & 181         & 191         & 198         & 191         & 157         \\ \hline
\textbf{Mask 20}                & 225         & 240         & 179         & 197         & 225         & 223         & 165         & 185         \\ \hline
\textbf{Mask 21}                & 353         & 315         & 318         & 295         & 300         & 321         & 268         & 243         \\ \hline
\textbf{Mask 22}                & 249         & 241         & 245         & 277         & 235         & 262         & 204         & 180         \\ \hline
\textbf{Mask 23}                & 211         & 254         & 200         & 246         & 236         & 238         & 188         & 172         \\ \hline
\textbf{Mask 24}                & 282         & 275         & 256         & 228         & 220         & 238         & 229         & 196         \\ \hline
\textbf{Mask 25}                & 277         & 239         & 258         & 252         & 241         & 242         & 201         & 224         \\ \hline
\textbf{Mask 26}                & 266         & 287         & 237         & 232         & 234         & 267         & 216         & 173         \\ \hline
\textbf{Mask 27}                & 213         & 177         & 208         & 154         & 182         & 204         & 120         & 105         \\ \hline
\textbf{Mask 28}                & 316         & 295         & 294         & 282         & 262         & 314         & 236         & 254         \\ \hline
\textbf{Mask 29}                & 278         & 329         & 321         & 300         & 269         & 387         & 274         & 247         \\ \hline
\textbf{Mask 30}                & 232         & 209         & 212         & 226         & 211         & 229         & 226         & 211         \\ \hline
\textbf{Mask 31}                & 254         & 265         & 237         & 189         & 234         & 255         & 178         & 181         \\ \hline
\textbf{Mask 32}                & 253         & 302         & 303         & 278         & 269         & 327         & 258         & 246         \\ \hline
\end{tabular}
\caption{Each column is a participant from the Natural Scenes Dataset. The first row represents the dimensionality of the fMRI inputs to the original decoding algorithm. Namely, we re-implemented a noise ceiling estimate based purely on our training set partition and included, per-participant, all voxels that had a noise ceiling greater than 5\%. The remaining columns relate to the dimensions of the sparse masks from Section 4.1 in the main text. }
\label{tab:voxel_counts}
\end{table}

We observe that the L1 regularization we apply in order to induce sparse masks results in similar voxel subset sizes for each shared decodable concept in $W_{SDC}$ irrespective of the number of voxels that passed the noise ceiling threshold and were used as inputs into the fMRI-decoding algorithm. Under our assumptions that a shared latent code across ROIs exists for shared decodable concepts, recovering broadly similar participant-specific and concept-specific mask voxel subset sizes is expected, although this does not itself mean they are similar in spatial distribution across the cortex. A beneficial feature of these masks is that they aren't limited to exact spatial overlap within ROI regions. This allows for participant-specific anatomical and functional specialization to be identified by our analysis method. We assess this using an ROI-similarity metric that checks for cross-participant consistency. 

\subsection{Visualization of Shared Decodable Concept Clusters via t-SNE}\label{tsne}
In order to explore the semantic meanings of each of the 32 dimensions in the shared decodable concepts matrix presented in Section 3.2. of the main text, we transform the CLIP representations of the full set of 73,000 stimulus images $\bm{Y}^{\text{Full}}_{\text{CLIP}} \in \mathbb{R}^{73000 \times 512}$ into a shared decodable concept space,  $SDC = \bm{Y}^{\text{Full}}_{\text{CLIP}} \bm{W}_{\text{SDC}}^T$. The top 10 highest-scoring images on each dimension are selected and plotted in Figure \ref{top_10_imgs}. To allow for a more thorough investigation of individual dimensions, we select the top 250 highest-scoring images and embed their CLIP vectors in a 2-dimensional space using the t-distributed stochastic neighbor embedding (t-SNE) method. A selection of t-SNE visualizations are plotted in Figures \ref{fig:tsne_animals}, \ref{fig:tsne_buildings}, \ref{fig:tsne_food}, \ref{fig:tsne_skies}, \ref{fig:tsne_toilets}. We plan to release the entire set of generated images and t-SNE projections upon manuscript acceptance. The t-SNE results in Section \ref{tsne} show remarkably coherent semantic networks containing hierarchical representations that cluster together in human-interpretable ways. We further explore the spatial organization of some of these dimensions by inspecting the participant-specific masks learned for these dimensions in order to derive hypotheses for future work out of our data-driven approach. 


\subsubsection{Food (Dimension 28)}
\begin{figure}
    \centering
    \includegraphics[width=15cm]{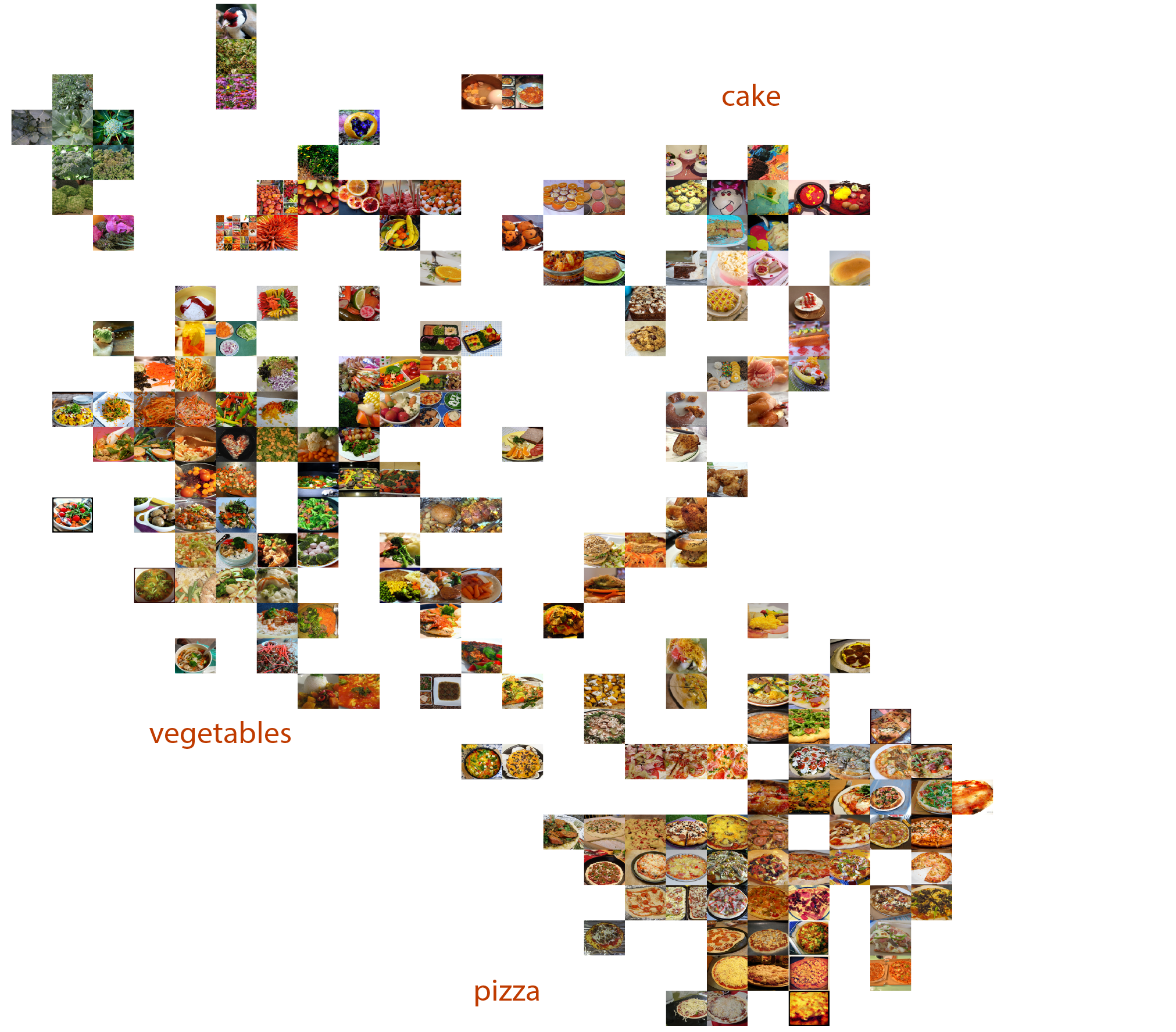}
    \caption{Dimension 28 represents the concept of food and spans many sub-clusters depending on food type, e.g. vegetables, pizza and cakes. We also observe a small cluster of leafy greens in the top-left. Recent work in probing the semantic representation of food in the ventral temporal cortex has implicated sensitivity to food as a core concept \citep{jain2023selectivity, kanwisher2022food}. We also find sensitivity to food as a shared decodable concept that is consistent across participants.}
    \label{fig:tsne_food}
\end{figure}

Figure \ref{fig:tsne_food} shows a 2D t-SNE projection of the top 250 images from this dimension into clusters.

\subsubsection{Skies (Dimension 19)}\label{app:skies}
\begin{figure}
    \centering
    \includegraphics[width=15cm]{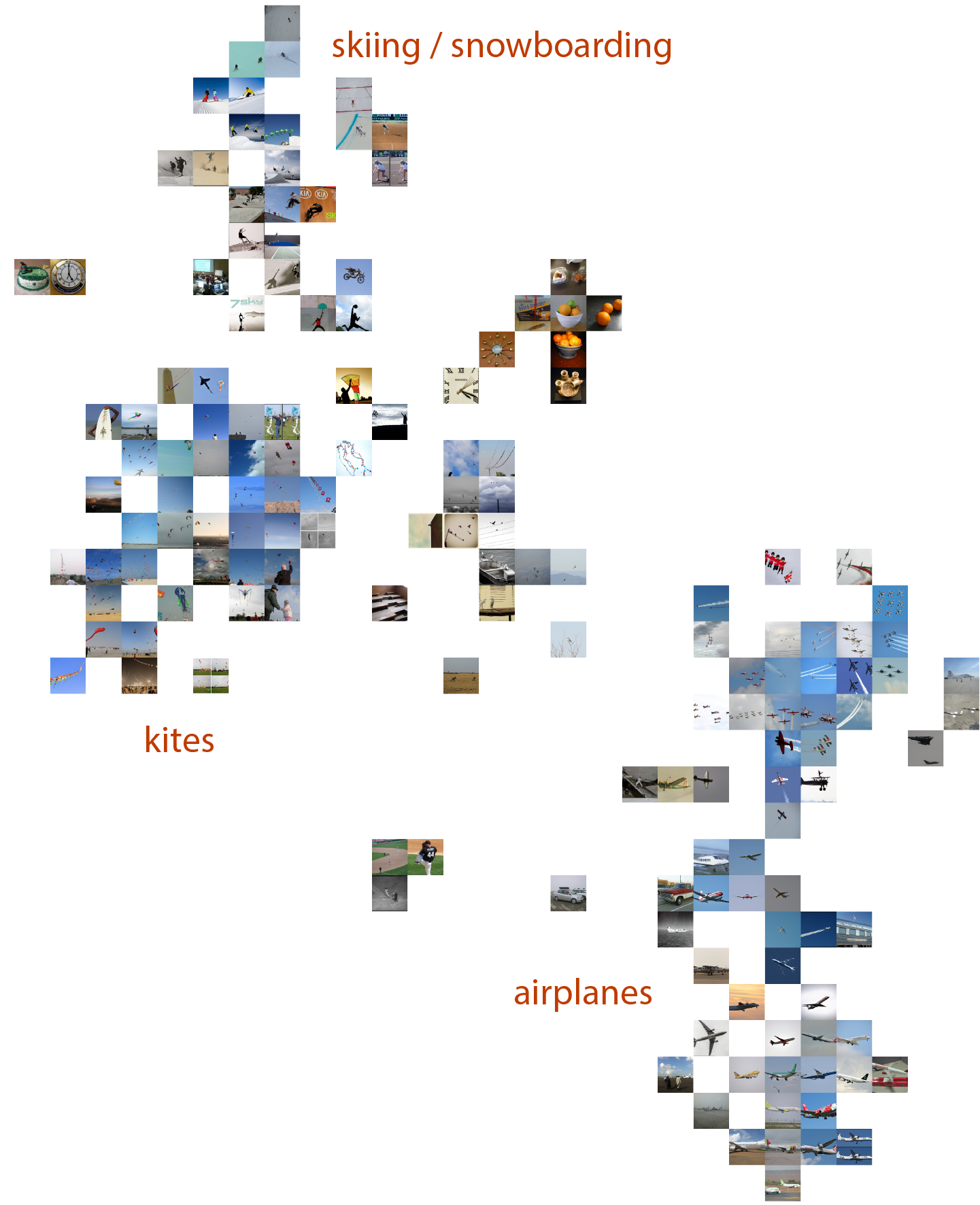}
    \caption{Dimension 19 appears to represent activity in the sky or on a consistent light background (i.e. ski slope). In this dimension we find clusters relating to airplane imagery, a different cluster specifically relating to flying kites, as well as one relating to skiing and snowboarding. In the skiing / snowboarding cluster, there are also images relating to skateboarding (without a strong sky background), a related activity. We find a few other images in this cluster that do not seem to be strongly related (clocks, sports, food).}
    \label{fig:tsne_skies}
\end{figure}

Figure \ref{fig:tsne_skies} shows a 2D t-SNE projection of the top 250 images from this dimension into clusters.

\subsubsection{Household Locations (Dimension 10)}
\begin{figure}
    \centering
    \includegraphics[width=15cm]{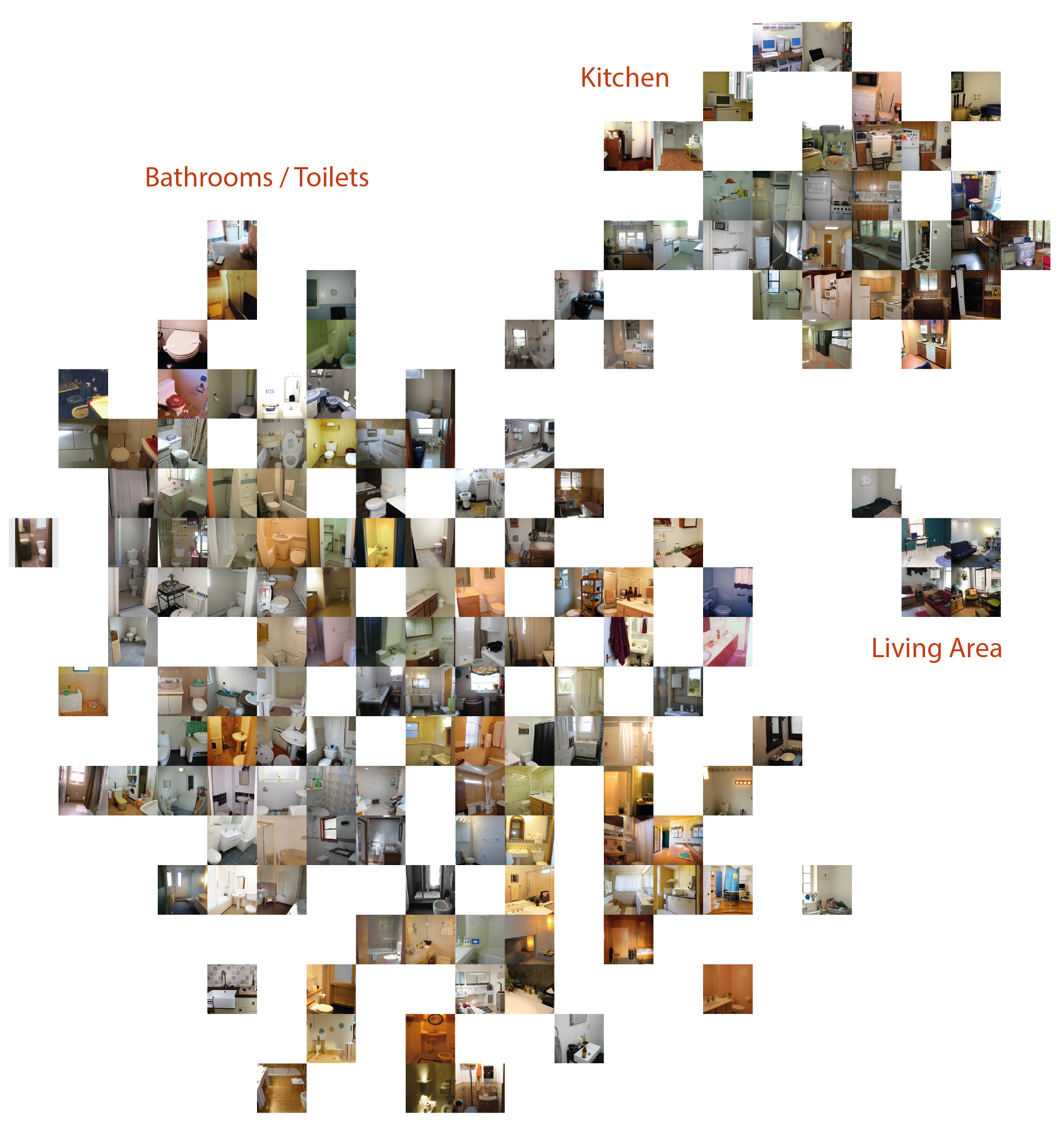}
    \caption{Dimension 10 contains two highly-related large clusters pertaining to bathrooms \& toilets (bottom left), and a primarily kitchen-related imagery in the top-right cluster. We also observe a smaller cluster (middle right) that appears to be more sensitive to living rooms. }
    \label{fig:tsne_toilets}
\end{figure}

Figure \ref{fig:tsne_toilets} shows a 2D t-SNE projection of the top 250 images from this dimension into clusters.

\subsubsection{Buildings (Dimension 26)}
\begin{figure}
    \centering
    \includegraphics[width=15cm]{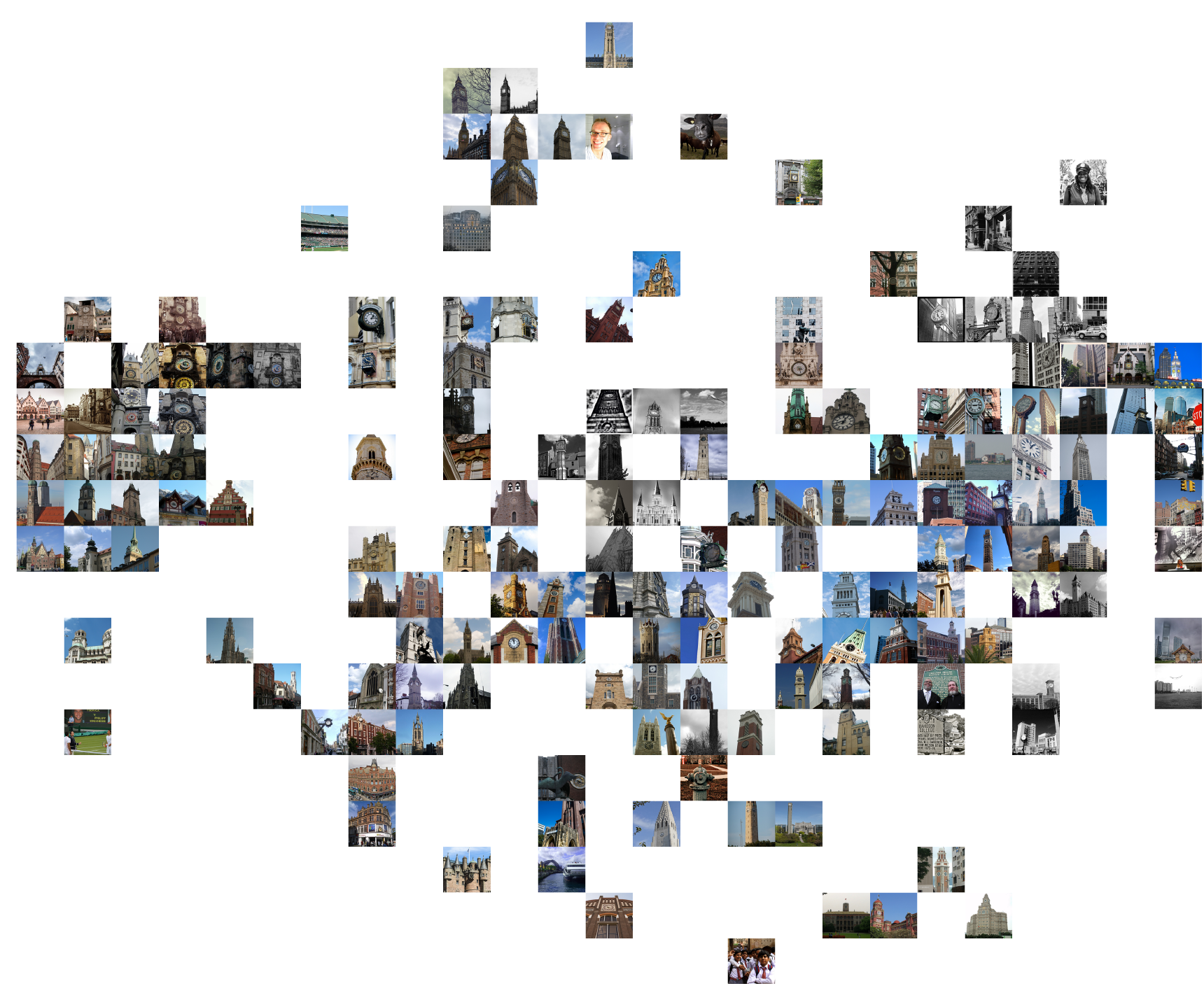}
    \caption{Dimension 26 encodes building representations. The most related images in CLIP space from the fMRI-decoded CLIP embeddings for this dimension are buildings in many different styles of image photography, in both black \& white and colour images and across a range of backgrounds.}
    \label{fig:tsne_buildings}
\end{figure}

Figure \ref{fig:tsne_buildings} shows a 2D t-SNE projection of the top 250 images from this dimension into clusters.

\subsection{Food}\label{app:food_flatmaps}

Recent findings have posited that areas encompassing and surrounding the PPA, FFA are highly selective for abstract food representations \citep{jain2023selectivity}, particularly the patch of cortex that separates them. After discovering that food was a shared decodable concept in our analysis, evidenced by the t-SNE clustering projection in Figure \ref{fig:tsne_food}, we explored our voxel masks across the 8 NSD participants in order to assess whether we would find similar results. 

\begin{figure}
    \centering
    \includegraphics[width=15cm]{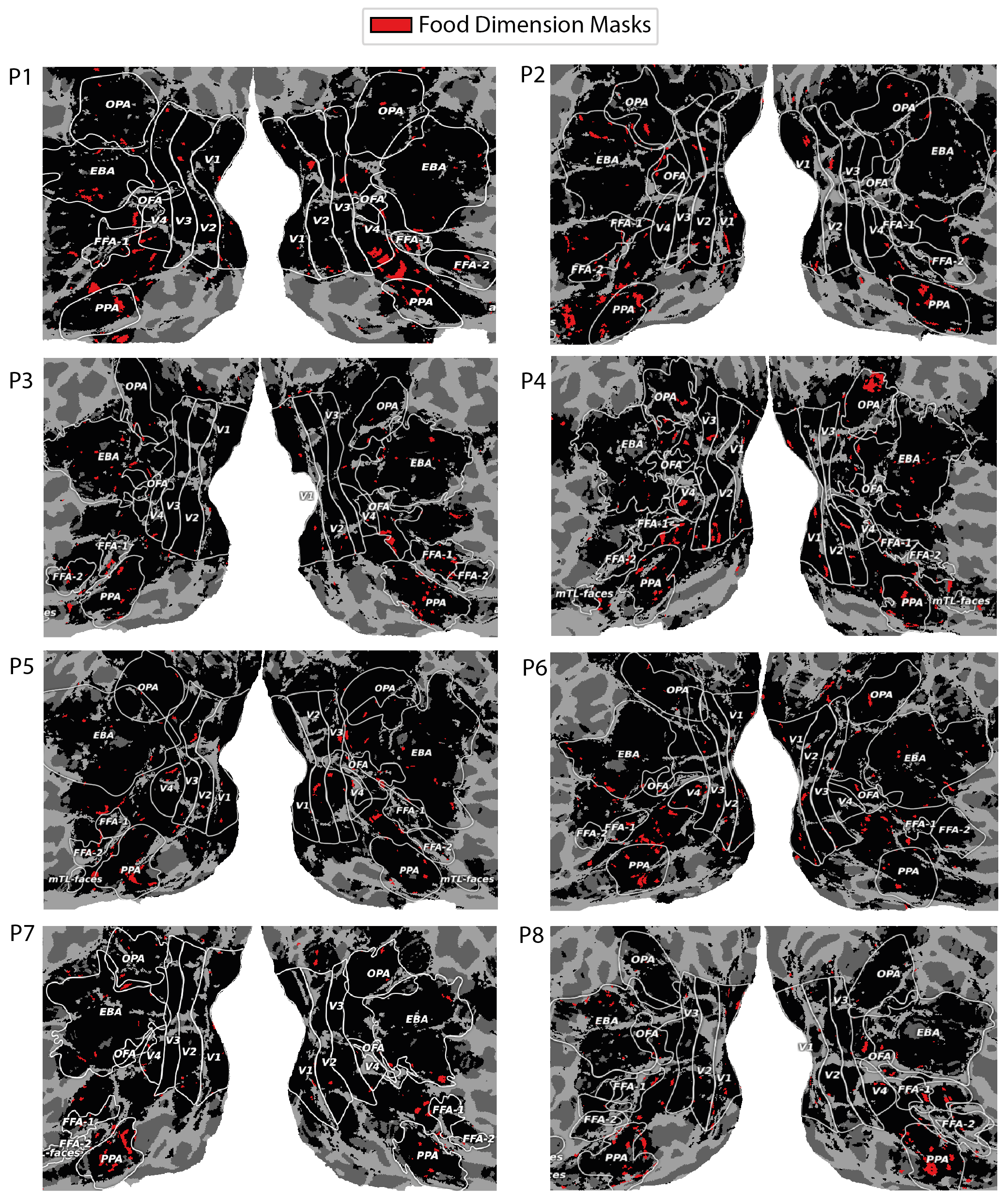}
    \caption{Flat maps of the 8 NSD participants and voxel masks corresponding to dimension 28, which we identified as being a dimension highly associated with food images. Prior research has linked the PPA, FFA and the area between them as being a functionally selective food region. We find our sparse masks lend some support to these results, particularly in Participant 1's results. }
    \label{fig:qual_food}
\end{figure}

Figure \ref{fig:qual_food} shows a varied set of results across participants. All participants have some level of voxel mask locations in the region between PPA and FFA (inclusive), while most do show a large presence in the regions previously identified as being specific for food. We don't restrict the area in which the voxel masks are learned and this could lead to some observed differences.

\end{document}